\documentclass[letterpaper]{article} 
\usepackage{aaai2027} 
\usepackage[hyphens]{url} 
\usepackage{graphicx} 
\usepackage{natbib} 
\usepackage{caption} 
\usepackage{booktabs}
\usepackage{amsfonts}
\usepackage{amsmath}
\usepackage{xcolor}
\usepackage{colortbl}
\usepackage{multirow}
\usepackage{longtable}
\usepackage{tabularx}
\usepackage{enumitem}
\usepackage{pifont}
\usepackage{dblfloatfix}

\providecommand{\HumanSubsetN}{150}

\providecommand{\HumanResponseN}{300}
\providecommand{\JudgeCalibrationN}{150}

\providecommand{\RefFullN}{124}
\providecommand{\RefFullPct}{82.7}
\providecommand{\RefPartialN}{23}
\providecommand{\RefPartialPct}{15.3}
\providecommand{\RefConflictN}{3}
\providecommand{\RefConflictPct}{2.0}
\providecommand{\RefUncertainN}{0}
\providecommand{\RefUncertainPct}{0.0}
\providecommand{\RefCompatiblePct}{98.0}  
\providecommand{\RefRevisedN}{3}
\providecommand{\RefRemovedN}{0}

\providecommand{\HumanBaselineSuccess}{83.0}
\providecommand{\HumanBaselineCILo}{77.7}
\providecommand{\HumanBaselineCIHi}{87.7}
\providecommand{\HumanBaselineV}{97.0}
\providecommand{\HumanBaselineI}{94.0}
\providecommand{\HumanBaselineK}{90.0}
\providecommand{\HumanBaselineR}{91.0}

\providecommand{\HumanHumanV}{0.79}
\providecommand{\HumanHumanI}{0.91}
\providecommand{\HumanHumanK}{0.85}
\providecommand{\HumanHumanR}{0.81}
\providecommand{\HumanHumanSuccess}{0.84}

\providecommand{\GeminiHumanV}{0.72}
\providecommand{\GeminiHumanI}{0.86}
\providecommand{\GeminiHumanK}{0.79}
\providecommand{\GeminiHumanR}{0.74}
\providecommand{\GeminiHumanSuccess}{0.78}
\providecommand{\GPTHumanV}{0.68}
\providecommand{\GPTHumanI}{0.83}
\providecommand{\GPTHumanK}{0.75}
\providecommand{\GPTHumanR}{0.70}
\providecommand{\GPTHumanSuccess}{0.74}
\providecommand{\IntersectionHumanV}{0.76}
\providecommand{\IntersectionHumanI}{0.88}
\providecommand{\IntersectionHumanK}{0.82}
\providecommand{\IntersectionHumanR}{0.77}
\providecommand{\IntersectionHumanSuccess}{0.81}

\definecolor{PromptHeader}{RGB}{229,236,243}
\definecolor{PromptCell}{RGB}{248,250,252}

\newenvironment{prompttable}[1]{%
  \par\medskip
  \begingroup
  \footnotesize
  \setlength{\tabcolsep}{5pt}
  \renewcommand{\arraystretch}{1.08}
  \noindent\begin{tabular}{>{\raggedright\arraybackslash}p{\dimexpr\columnwidth-2\tabcolsep\relax}}
  \toprule
  \rowcolor{PromptHeader}\sffamily\bfseries #1\tabularnewline
  \midrule
  \rowcolor{PromptCell}
}{%
  \tabularnewline
  \bottomrule
  \end{tabular}
  \endgroup
  \par\smallskip
}

\title{MemeBench: What LVLMs Miss When Interpreting Culture-Dependent Memes}

\author{
  Weihang Wang\textsuperscript{\rm 1,\rm 2},
  Kainan Tu\textsuperscript{\rm 2},
  Jielei Zhang\textsuperscript{\rm 1}\thanks{Project leader.},
  Run Yang\textsuperscript{\rm 1},
  Boheng Sheng\textsuperscript{\rm 1},
  Yuchen He\textsuperscript{\rm 1},
  \\
  Yu Xie\textsuperscript{\rm 1},
  Pengyu Chen\textsuperscript{\rm 1},
  Peiyi Li\textsuperscript{\rm 1},
  Huyang Sun\textsuperscript{\rm 1},
  Longwen Gao\textsuperscript{\rm 1},
  Zhouhui Lian\textsuperscript{\rm 3}\corresponding
}

\affiliations{
  \textsuperscript{\rm 1}Bilibili\\
  \textsuperscript{\rm 2}Fudan University\\
  \textsuperscript{\rm 3}Peking University\\
  \{kiren.wwh, yctmzjl\}@gmail.com
}

\nocopyright

\begin{document}

\maketitle

\begin{abstract}
Large vision-language models have improved at describing
visual content, but accurate descriptions do not ensure
interpretation when meaning depends on knowledge beyond the pixels.
Memes expose this gap because they rely on cultural entities,
background knowledge, and community conventions.
Most meme benchmarks reduce interpretation to labels or holistic
scores, obscuring where an explanation breaks
down.
We introduce \textbf{MemeBench}, a diagnostic benchmark of 1,253
Chinese and English memes with human-written references and
quality-controlled VIKR annotations, centered on anime, comics, games,
and adjacent online subcultures.
Its \textbf{VIKR} schema decomposes explanations into Visual clues,
Identity links, Knowledge units, and Reasoning mechanisms.
Across 26 LVLMs, every model covers visible content more reliably than
the knowledge needed to interpret it, and even the strongest retains a
22.6\% Visual--Knowledge gap.
To test whether this diagnosis can guide improvement, we introduce
\textbf{KAR}, an entity-guided retrieval baseline built on
\textbf{CultureBase}.
Across four controlled models, KAR raises VIKR Success by
3.6--7.4\% and, compared with generic retrieval,
repairs more answers and breaks fewer.
Yet both retrieval conditions improve Identity and Knowledge while
reducing Visual coverage in every comparison.
MemeBench reveals whether an interpretation succeeds, what is missing,
and whether targeted evidence fills the diagnosed gap.
\end{abstract}

\begin{figure*}[t]
\centering
\includegraphics[width=\textwidth]{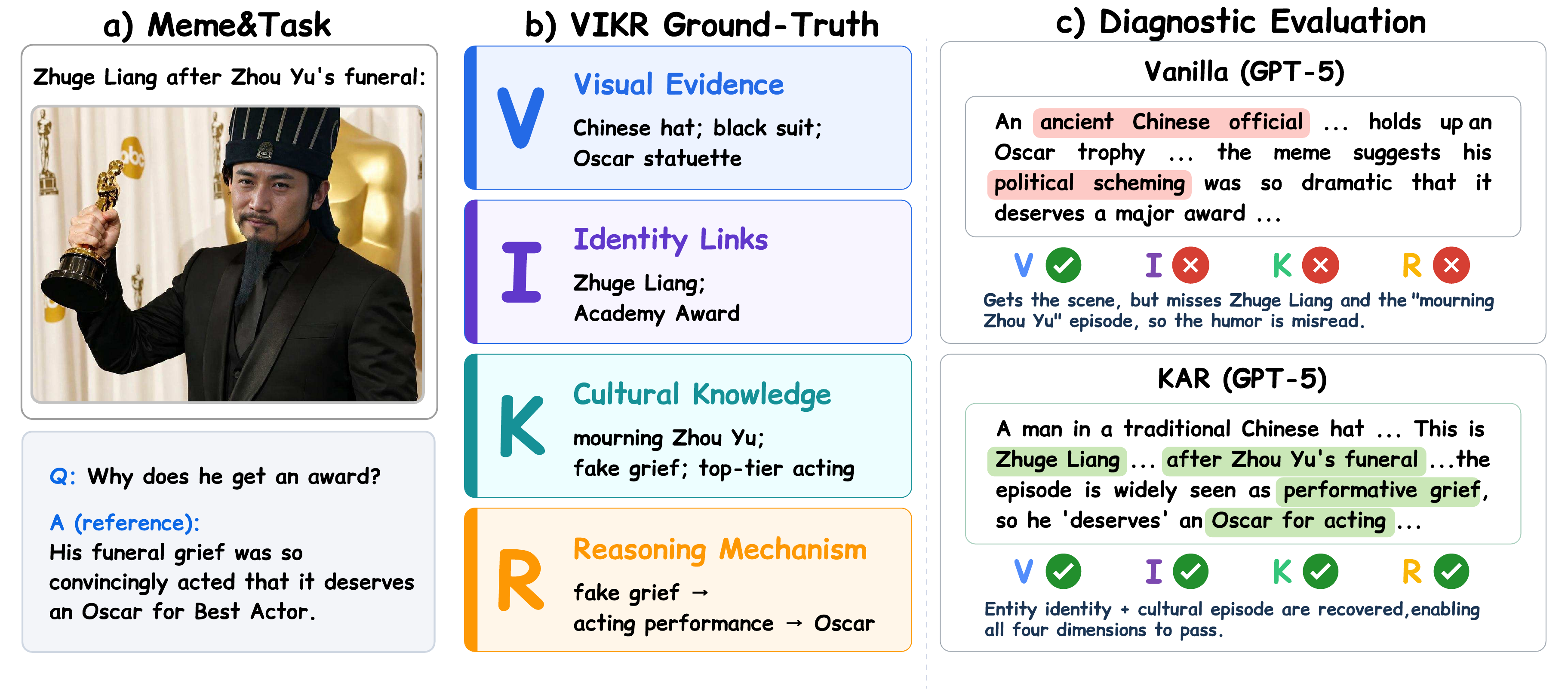}
\caption{{The task and evaluation of MemeBench. \textbf{a)} meme with reference answer. \textbf{b)} VIKR ground truth and per-dimension checklists. \textbf{c)} In this paired example, the Vanilla response misses the entity and the cultural episode and fails I/K/R; KAR (\S\ref{sec:kar_method}) retrieves both and passes all four. More examples in Appendix~\ref{app:vikr_case_table}.}}
\label{fig:kar_case_study}
\end{figure*}

\section{Introduction}
\label{sec:intro}

Large vision-language models (LVLMs) have made rapid progress in recognizing and describing visual content.
Yet an accurate description does not guarantee a correct interpretation when meaning depends on knowledge beyond the pixels.
In Figure~\ref{fig:kar_case_study}, an LVLM correctly describes an ancient Chinese official holding an Oscar statuette, but fails to recognize Zhuge Liang mourning Zhou Yu and interprets the scene as political scheming rather than performative grief.
As LVLMs are increasingly expected to interpret open-world visual content, measuring this separation between seeing and meaning becomes more important~\citep{cohen2025entitygap,yadav2025cultural,chiu2025culturalbench,myung2024blend}.
\noindent\begin{minipage}{\columnwidth}
\centering
\scriptsize
\setlength{\tabcolsep}{1.2pt}
\begin{tabular*}{\columnwidth}{@{\extracolsep{\fill}}lccccc@{}}
\toprule
\textbf{Benchmark} & \textbf{Explanation} & \textbf{Bilingual} & \textbf{Decomposition} & \textbf{Diagnosis} & \textbf{Augmentation} \\
\midrule
Hateful Memes & \ding{55} & \ding{55} & \ding{55} & \ding{55} & \ding{55} \\
MemeCap & \ding{51} & \ding{55} & \ding{51} & \ding{55} & \ding{55} \\
D-HUMOR & \ding{55} & \ding{55} & \ding{55} & \ding{55} & \ding{55} \\
MemeIntent & \ding{51} & \ding{55} & \ding{51} & \ding{55} & \ding{55} \\
MUnd/MetaGPT & \ding{51} & \ding{51} & \ding{51} & \ding{55} & \ding{55} \\
MemeLens & \ding{55} & \ding{51} & \ding{55} & \ding{55} & \ding{55} \\
M-QUEST & \ding{55} & \ding{55} & \ding{51} & \ding{51} & \ding{55} \\
MemeQA & \ding{55} & \ding{55} & \ding{51} & \ding{51} & \ding{55} \\
CII-Bench & \ding{55} & \ding{55} & \ding{55} & \ding{55} & \ding{55} \\
\midrule
\textbf{MemeBench} & \textbf{\ding{51}} & \textbf{\ding{51}} & \textbf{\ding{51}} & \textbf{\ding{51}} & \textbf{\ding{51}} \\
\bottomrule
\end{tabular*}
\captionof{table}{Comparison with existing meme benchmarks: open-ended explanation rather than labels or QA (Explanation), Chinese and English coverage (Bilingual), annotation of intermediate meaning components (Decomposition), component-level scoring of responses (Diagnosis), and supplying targeted evidence to test component-level changes (Augmentation).}
\label{tab:comparison}
\end{minipage}\par\medskip

Existing benchmarks provide limited resolution for studying this problem.
General-purpose LVLM benchmarks assess broad visual perception and reasoning~\citep{liu2024mmbench,yue2024mmmu,yu2024mmvet,fu2023mme}, while meme benchmarks have expanded from classification to captioning, question answering, and open-ended explanation~\citep{nguyen2025memeqa,degiorgis2026mquest,zhou2026figurative,zhao2025memereacon,zhong2024fime}.
Most evaluations nevertheless reduce an interpretation to a label, an answer, or a holistic correctness score.
Such a score can indicate whether a response is acceptable, but cannot show whether it misdescribes the image, misses the referenced entity, lacks the relevant background knowledge, or fails to connect these elements into the intended meaning.
Models with similar overall scores may omit different information, and the same augmentation may improve one part of an explanation while weakening another.
Table~\ref{tab:comparison} compares MemeBench with existing benchmarks along these dimensions.

We introduce \textbf{MemeBench}, a bilingual diagnostic benchmark for open-ended meme interpretation.
It contains 1{,}253 Chinese and English memes collected from Bilibili, Reddit, and ImgFlip, centered on ACG (anime, comics, and games) and adjacent online subcultures.
These communities provide a concentrated test bed for knowledge-dependent image interpretation because their memes often rely on long-tail characters, source works, community events, and conventions.
MemeBench references 2{,}072 distinct entities, most of which occur in only one meme.
Unlike recognition or multiple-choice evaluation, open-ended interpretation requires models to supply rather than select the relevant entities and cultural background.
This exposes incomplete but plausible explanations and tests whether external evidence recovers missing content without displacing correct content.

At the core of MemeBench is \textbf{VIKR}, which organizes a complete explanation into four kinds of observable content.
They cover \textbf{V}isual clues, \textbf{I}dentity links, \textbf{K}nowledge units, and \textbf{R}easoning mechanisms.
Binary checklists measure which requirements a response covers within each dimension.
Passing every checklist in all four dimensions marks a complete interpretation, while individual failures localize omitted content.
VIKR characterizes the observable semantics of an explanation while leaving model-internal processing unspecified.
It can identify what an explanation is missing and test whether an augmentation fills the diagnosed gap.

Our evaluation of 26 commercial and open-source LVLMs reveals a consistent separation between visual description and culturally grounded interpretation.
Every model covers visible content more reliably than the background knowledge needed to explain it.
This separation remains substantial even for the strongest model, whose Visual coverage exceeds its Knowledge coverage by 22.6\%.
VIKR also distinguishes models that a single \textsc{Success} score places together, showing that similar aggregate performance can conceal different missing content.
The two language-associated online-cultural ecosystems exhibit the same broad structure, with their larger differences concentrated in Identity and Knowledge rather than Visual coverage.

We next ask whether this diagnosis can guide improvement.
We compare reasoning guidance, generic web retrieval, and \textbf{KAR}, an entity-guided retrieval condition that grounds cultural entities through \textbf{CultureBase} before web search.
Reasoning guidance alone does not produce a consistent improvement.
Across the four controlled models, both retrieval conditions improve Identity and Knowledge, while Visual coverage declines in all eight retrieval comparisons.
In paired comparison with generic web retrieval, KAR further improves joint \textsc{Success} for every model, repairs more answers, and breaks fewer.
A positive net score can conceal opposing changes within an explanation.
By revealing what an explanation gets right, what it misses, and how retrieval changes both, MemeBench turns open-world meme interpretation from a single score into an actionable diagnosis for model improvement.
This capability matters when culturally specific visual content must be interpreted across communities.

Our contributions are threefold.
\begin{enumerate}[nosep,leftmargin=*]
  \item We introduce \textbf{MemeBench}, a bilingual benchmark of 1{,}253 Chinese and English memes, with VIKR annotations and dimension-specific checklists for diagnosing open-ended interpretations.
  \item Through an evaluation of 26 LVLMs, we show that models cover visible content more reliably than the entities and background knowledge needed for interpretation, while similar overall scores can conceal different missing content.
  \item We introduce \textbf{KAR} as an entity-guided retrieval baseline and show that targeted evidence changes explanation components selectively, demonstrating how MemeBench evaluates both component-level gains and preservation under model augmentation.
\end{enumerate}

\newpage
\section{Related Work}
\label{sec:related}

\paragraph{Meme understanding benchmarks.}
Meme benchmarks have progressed from harmful-content classification~\citep{kiela2020hateful,pramanick2021momenta,bui2024multi3hate,lu2024toxicnmm} to captioning~\citep{hwang2023memecap} and higher-level semantics~\citep{kasu2025dhumor,nandy2024yesbut,zhang2025ciibench,xu2025punmemecn}.
MemeIntent~\citep{park2024memeintent} generates intent descriptions with annotated background knowledge.
MetaGPT~\citep{xu2026metagpt} targets metaphor detection, domain extraction, and interpretation, while MemeLens~\citep{shahroor2026memelens} specializes a multilingual multitask VLM for memes.
Cross-cultural work studies transcreation and hateful-meme detection~\citep{zhao2026transcreation,wang2026nativememes}, while MemeQA~\citep{nguyen2025memeqa} and M-QUEST~\citep{degiorgis2026mquest} decompose understanding through QA.
These tasks broaden coverage but do not diagnose free-form explanations through requirement-level Visual, Identity, Knowledge, and Reasoning coverage.

\paragraph{Knowledge-augmented multimodal models.}
CulturalVQA~\citep{nayak2024culturalvqa} probes geographically grounded cultural knowledge through visual QA, while BLEnD-Vis~\citep{tan2026blendvis} targets everyday cultural knowledge through visual QA and multiple-choice variants.
AVMeme Exam~\citep{jiang2026avmeme} extends cultural evaluation to audio-visual memes.
Retrieval-augmented generation has been extended to visual QA~\citep{lewis2020rag,chen2022murag,caffagni2024wikillava,li2024searchlvlms}, with entity-centric resources supporting factual QA and recognition~\citep{marino2019okvqa,chen2023infoseek,mensink2023encyclopedic,hu2023oven}.
These settings do not connect requirement-level diagnosis to retrieval: search agents learn \emph{when} to search~\citep{wu2025mmsearchr1,zhang2026vsearcher}, whereas KAR uses entity grounding to determine \emph{what} cultural knowledge to query before issuing web searches.

\medskip
\noindent\begin{minipage}{\columnwidth}
  \centering
  \includegraphics[width=\columnwidth]{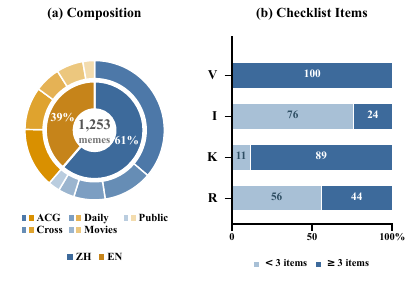}
  \captionof{figure}{Benchmark scope and evaluation granularity. \textbf{(a)} Language (inner ring) and domain composition within each language (outer ring); per-cell counts in Appendix~\ref{app:selection}. \textbf{(b)} Checklist density by dimension, shown as the share of memes with fewer than three versus at least three items. A meme carries 12.2 items on average; \textsc{Success} requires all items to pass.}
  \label{fig:benchmark_stats}
\end{minipage}

\newpage

\section{MemeBench}
\label{sec:benchmark}
\label{sec:task}

MemeBench evaluates open-ended meme interpretation.
Given a meme image and a standard explanation prompt, a model produces a free-form explanation without candidate answers or multiple-choice options.
Evaluation uses the per-dimension VIKR checklists to measure coverage of visible content, invoked entities, relevant background, and the reasoning that connects them (Figure~\ref{fig:kar_case_study}).
Because an answer may satisfy only some requirements, these scores preserve both covered and omitted content instead of collapsing the explanation into a holistic judgment.

\subsection{Data Construction}
\label{sec:data}

MemeBench centers on ACG and adjacent online cultures as a concentrated stress test for interpretation that depends on long-tail entities and community-specific knowledge.
It comprises \textbf{1,253 memes} (768 Chinese, 485 English) collected between June and December 2025: Chinese samples come from Bilibili, a platform centered on ACG and internet culture, while English samples come from Reddit and ImgFlip.

Construction begins with manual screening for three properties: cultural knowledge beyond surface recognition, an explicitly explainable intended meaning, and sufficient image quality, yielding \textbf{1,500 candidate memes}.
For each retained candidate, a bilingual annotator familiar with ACG and internet meme culture writes a reference explanation covering visual content, entity identity, cultural background, and humor mechanism, and assigns language and domain metadata.
These human-written explanations serve as the semantic source for subsequent structured annotation.
Gemini-3.1-Pro then converts each explanation into the VIKR schema (\S\ref{sec:annotation}).
Quality control combines automated schema and entity-link validation with targeted manual review of VIKR layer separation, checklist consistency, ambiguity, image quality, and safety; 247 items are removed, producing the final \textbf{1,253-item} evaluation set.
A reference-independence study on a language--domain-stratified subset of \HumanSubsetN{} items asks two participants to re-explain each meme under blind conditions, with conflicts adjudicated independently.
The study finds \RefCompatiblePct\% of references fully or partially compatible and leads to \RefRevisedN{} revisions and \RefRemovedN{} removals; Appendix~\ref{app:human_study} gives the protocol and Appendix~\ref{app:selection} the full selection breakdown.

\subsection{VIKR Annotation Schema}
\label{sec:annotation}

VIKR represents a complete explanation through four observable response requirements: \textbf{V}isual clues, \textbf{I}dentity links, \textbf{K}nowledge units, and \textbf{R}easoning mechanisms.
This representation turns free-form evaluation into a structured diagnosis that localizes missing content while keeping every requirement tied to the same reference interpretation.
The schema operates entirely at the response level.
It distinguishes whether a response describes the scene, grounds the referenced entities, supplies the required cultural background, and connects these elements to the intended meaning.
These requirements may co-occur in an explanation, but each is evaluated independently rather than used as a proxy for another.

The \textbf{Visual~(V)} layer provides an entity-agnostic visual description, including OCR transcription and the appearance of salient entities, without naming them.
The \textbf{Identity~(I)} layer names entities and their source works, each linked by a shared ID to its appearance in the V~layer.
Separating Identity from Knowledge is diagnostically important because recognizing an entity does not determine which of its many cultural associations the scene invokes.
The \textbf{Knowledge~(K)} layer captures the background facts and cultural codes the meme requires, and the \textbf{Reasoning~(R)} layer documents the humor trigger, mechanism, and core message.
Each meme also carries language and domain metadata for stratified analysis; metadata does not participate in evaluation.

\subsection{Evaluation Protocol}
\label{sec:eval}

Each VIKR layer yields 1--4 binary checklist items, and a layer's score $d_i \in \{0,1\}$ for $d \in \{V,I,K,R\}$ is set to 1 only if all its items are satisfied.
The Identity checklist requires exact entity identification (``K\={e}n\'{a}n'' [Conan] rather than ``a detective boy''); the Reasoning checklist accepts semantically equivalent formulations.
This item-level representation supports both dimension-specific diagnosis and a strict all-dimension primary metric:
\begin{equation}
  \mathrm{Success}_i = V_i I_i K_i R_i,
  \label{eq:success}
\end{equation}
which is 1 exactly when the response satisfies every annotated requirement.
The product preserves the four diagnostic scores while reserving overall success for explanations that cover the complete interpretation.
Appendix~\ref{app:metric_sensitivity} reports a three-of-four sensitivity analysis.

Gemini-3.1-Pro and GPT-5.1 independently score each response against the same structured checklist (Appendix~\ref{app:prompt}), seeing the meme and response but not the model identity.
Following the LLM-as-a-judge paradigm~\citep{zheng2023judging,lee2025checkeval}, an item passes only when both judges agree.
Dimension scores aggregate from these intersection decisions.
On the independently human-labeled calibration subset, this intersection reaches Cohen's $\kappa{=}\IntersectionHumanSuccess$ on \textsc{Success}, above either judge alone ($\GeminiHumanSuccess$ and $\GPTHumanSuccess$) and within $0.03$ of the $\HumanHumanSuccess$ agreement between two human raters (Appendix~\ref{app:human_study}).
Across all 26 models, the single-judge leaderboards also agree at Spearman $\rho{=}0.99$, showing that the judges differ mainly in strictness rather than model ranking (Appendix~\ref{app:judge_robustness}).

\begin{figure*}[t]
  \centering
  \includegraphics[width=0.96\textwidth]{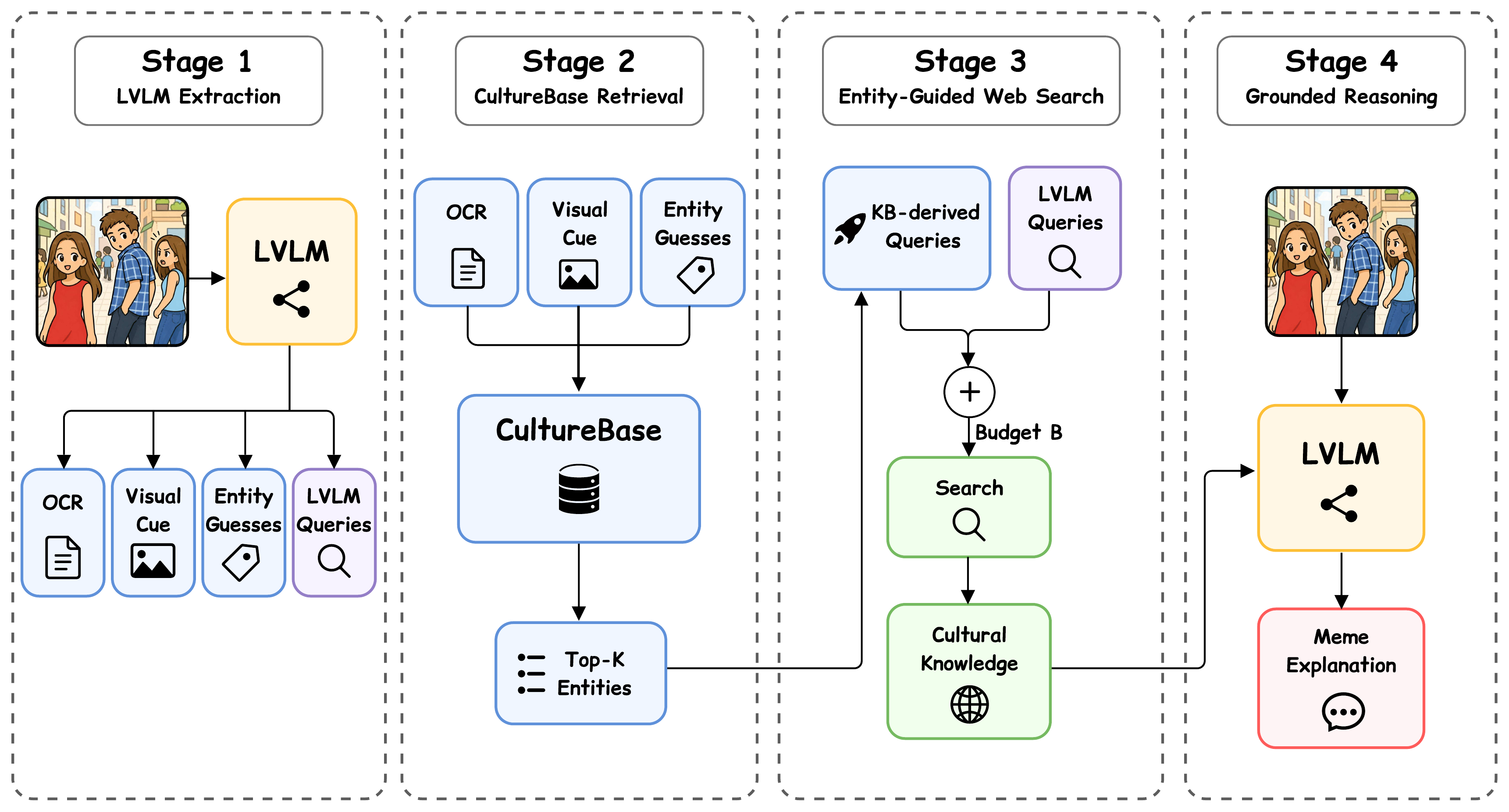}
  \caption{\textbf{KAR pipeline.} LVLM cue and entity-hypothesis extraction, CultureBase grounding, budgeted entity-guided web search, and evidence-grounded explanation.}
  \label{fig:kar}
\end{figure*}

\subsection{Benchmark Statistics}

MemeBench contains 628 ACG items; the remaining 625 span cross-domain references, daily life, movies and television, and public life, broadening the range of cultural references within this sampling frame (Figure~\ref{fig:benchmark_stats}(a); counts in Appendix~\ref{app:selection}).
Because the Chinese and English subsets come from different platforms and domain mixtures, cross-language results characterize two language-associated online-cultural ecosystems.
The benchmark references 2,072 distinct entities, 86.9\% of which appear in exactly one meme, while even the most frequent entity occurs in only 1.3\% of items.
A further 27.6\% of memes cite at least two source works, so coverage of a few popular characters is insufficient.
Its annotations contain 15,298 binary checklist items, 12.2 per meme on average (Figure~\ref{fig:benchmark_stats}(b)).
Visual and Knowledge contain at least three items for 99.8\% and 88.6\% of memes, respectively, and \textsc{Success} requires every annotated item to pass.
Together, the long-tailed reference space and dense checklist requirements test both the breadth of cultural grounding and the completeness of each explanation.

\section{Entity-Guided Retrieval}
\label{sec:kar_method}

To test whether the diagnosed bottlenecks respond to targeted evidence, we build \textbf{KAR} (\textbf{K}nowledge-\textbf{A}ware \textbf{R}etrieval), an entity-guided retrieval condition that grounds cultural entities before searching the web.
The design uses entity identity as the handle by which background knowledge becomes searchable, since a query anchored to the right entity reaches background that a generic visual description cannot.

\paragraph{CultureBase.}
The grounding step draws on \textbf{CultureBase}, a curated multilingual inventory of cultural entities.
Each record stores the canonical name, aliases, source work, character traits, visual description, and cultural usage patterns of one entity, embedded with bge-m3~\citep{chen2024bge} for multilingual similarity retrieval.
CultureBase supplies entity anchors and query context; the final explanation is generated jointly from the original image and retrieved web evidence.

\paragraph{Pipeline.}
KAR runs in four stages, shown in Figure~\ref{fig:kar}.
Stage~1 prompts the LVLM to extract OCR text $o_i$, a short visual cue phrase $h_i$, entity hypotheses $\mathcal{H}_i$ with confidence levels, and model-suggested web queries $\mathcal{Q}^{\mathrm{vlm}}_i$.
Stage~2 forms a three-route query set from the OCR text, cue phrase, and medium- or high-confidence entity names:
\begin{equation}
\mathcal{R}_i =
  \{o_i,h_i\}\cup
  \{n_\ell\mid c_\ell\in\{\mathrm{medium},\mathrm{high}\}\}.
\end{equation}
Each CultureBase entity is scored by its best match across these routes:
\begin{equation}
  \mathrm{score}_{ij}=\max_{r\in\mathcal{R}_i} g(r)^\top u_j,
\end{equation}
where $g(\cdot)$ is the bge-m3 embedding function and $u_j$ is the embedding of entity $e_j$.
KAR retains the top five entities above the fixed similarity threshold $\tau{=}0.5$, allowing textual and visual cues to recover an identity even when the model's initial name hypothesis is incomplete.
Stage~3 converts each retained entity into a templated query containing its name, source, and meme origin, then merges these queries with $\mathcal{Q}^{\mathrm{vlm}}_i$ under a shared budget.
The merged set is issued to Tavily, and Stage~4 returns the retrieved evidence with the original image for the final explanation.

The generic condition, Search-CoT, shares the extractor, search interface, query budget, evidence formatting, and grounded reasoner, and differs only in omitting the CultureBase stage.
If no CultureBase entity exceeds $\tau$, no entity-grounded query is added and KAR follows the same query path as Search-CoT.
Both retrieval conditions use two LVLM calls and at most five Tavily queries per item; KAR adds one local CultureBase lookup while keeping the web-query budget fixed.
This isolates entity grounding under a fixed web backend and evidence budget; Appendix~\ref{app:kardetails} gives prompts, hyperparameters, and coverage definitions.

\begin{table*}[t]
\centering
\scriptsize
\setlength{\tabcolsep}{1.2pt}
\setlength{\aboverulesep}{0.25ex}
\setlength{\belowrulesep}{0.25ex}
\renewcommand{\arraystretch}{0.84}
\captionsetup{skip=3pt}
\begin{tabular*}{\textwidth}{@{\extracolsep{\fill}}lrrrrrrrrrrrrrrr@{}}
\toprule
\textbf{Model} & \multicolumn{5}{c}{\textbf{Overall}} & \multicolumn{5}{c}{\textbf{Chinese}} & \multicolumn{5}{c}{\textbf{English}} \\
\cmidrule(lr){2-6}\cmidrule(lr){7-11}\cmidrule(l){12-16}
& \textsc{Succ.} & \textbf{V} & \textbf{I} & \textbf{K} & \textbf{R} & \textsc{Succ.} & \textbf{V} & \textbf{I} & \textbf{K} & \textbf{R} & \textsc{Succ.} & \textbf{V} & \textbf{I} & \textbf{K} & \textbf{R} \\
\midrule
\multicolumn{16}{c}{\emph{Commercial Models}} \\[-0.5pt]
Gemini-3.1-Pro & \textbf{60.3} & \textbf{92.3} & \textbf{74.6} & \textbf{69.7} & \textbf{76.7} & \textbf{49.9} & \textbf{90.4} & \textbf{66.8} & \textbf{61.1} & \textbf{67.6} & \textbf{76.7} & \textbf{95.3} & \textbf{87.0} & \textbf{83.3} & \textbf{91.1} \\
Gemini-3-Flash & 53.7 & 86.0 & 68.7 & 65.5 & 70.3 & 41.8 & 82.2 & 58.9 & 54.8 & 59.4 & 72.6 & 92.2 & 84.3 & 82.5 & 87.6 \\
GPT-5.6-Sol & 40.9 & 72.5 & 58.7 & 55.1 & 60.7 & 33.1 & 69.3 & 51.6 & 46.4 & 51.3 & 53.4 & 77.7 & 69.9 & 69.1 & 75.7 \\
Claude-Opus-5 & 38.3 & 74.7 & 50.4 & 50.9 & 57.9 & 28.3 & 69.9 & 41.5 & 39.7 & 46.9 & 54.2 & 82.3 & 64.5 & 68.7 & 75.5 \\
Qwen3.6-Plus & 28.4 & 65.1 & 47.2 & 42.8 & 47.7 & 25.5 & 62.9 & 44.0 & 37.8 & 41.0 & 33.0 & 68.7 & 52.4 & 50.7 & 58.4 \\
GPT-5.6-Terra & 27.8 & 61.9 & 45.4 & 42.0 & 46.9 & 22.3 & 57.2 & 40.5 & 33.2 & 37.8 & 36.5 & 69.5 & 53.2 & 55.9 & 61.4 \\
GPT-5.6-Luna & 18.0 & 52.1 & 34.9 & 31.3 & 35.0 & 15.5 & 49.1 & 31.4 & 24.9 & 28.8 & 22.1 & 56.9 & 40.4 & 41.4 & 44.7 \\
Claude-Sonnet-5 & 17.8 & 59.2 & 33.3 & 30.2 & 35.6 & 12.2 & 55.3 & 25.0 & 20.8 & 26.0 & 26.6 & 65.4 & 46.4 & 44.9 & 50.7 \\
Qwen3.6-Flash & 12.3 & 53.2 & 28.6 & 25.7 & 29.0 & 10.5 & 50.7 & 26.2 & 21.0 & 22.7 & 15.1 & 57.1 & 32.4 & 33.2 & 39.0 \\
\midrule
\multicolumn{16}{c}{\emph{Open-Source Models}} \\[-0.5pt]
Kimi-K2.5 & 37.9 & 71.1 & 59.2 & 54.5 & 58.7 & 32.4 & 68.1 & 56.1 & 48.2 & 50.5 & 46.6 & 75.9 & 64.1 & 64.5 & 71.5 \\
Qwen3-VL-235B-A22B-Instruct & 14.5 & 53.2 & 30.6 & 26.1 & 27.3 & 10.2 & 49.9 & 26.0 & 18.9 & 20.3 & 21.4 & 58.4 & 37.9 & 37.5 & 38.4 \\
MiMo-V2-Omni & 14.3 & 48.1 & 29.7 & 26.4 & 28.8 & 10.8 & 44.3 & 24.9 & 19.4 & 21.2 & 19.8 & 54.2 & 37.3 & 37.5 & 40.8 \\
MiMo-V2.5 & 13.7 & 47.2 & 30.5 & 26.9 & 28.3 & 9.9 & 43.6 & 25.7 & 19.5 & 19.4 & 19.8 & 52.8 & 38.1 & 38.6 & 42.5 \\
Qwen3-VL-235B-A22B-Thinking & 12.0 & 50.2 & 27.5 & 22.9 & 24.7 & 9.6 & 48.4 & 23.0 & 17.6 & 19.3 & 15.7 & 53.0 & 34.6 & 31.3 & 33.4 \\
Qwen3-VL-32B-Instruct & 8.4 & 47.6 & 20.1 & 19.1 & 20.0 & 5.9 & 44.4 & 16.5 & 13.9 & 13.2 & 12.4 & 52.8 & 25.8 & 27.2 & 30.9 \\
Qwen3-VL-32B-Thinking & 7.7 & 44.5 & 20.5 & 17.9 & 19.7 & 5.5 & 41.5 & 17.1 & 12.5 & 13.4 & 11.1 & 49.3 & 26.0 & 26.4 & 29.7 \\
Qwen3-VL-30B-A3B-Thinking & 5.8 & 41.6 & 18.3 & 14.7 & 15.8 & 3.6 & 38.0 & 14.1 & 10.5 & 10.7 & 9.3 & 47.2 & 24.9 & 21.2 & 23.9 \\
Qwen3-VL-8B-Thinking & 3.4 & 32.1 & 14.0 & 10.0 & 9.9 & 2.7 & 30.5 & 12.0 & 6.6 & 7.8 & 4.3 & 34.6 & 17.3 & 15.3 & 13.2 \\
Qwen3-VL-30B-A3B-Instruct & 3.3 & 39.7 & 14.8 & 12.2 & 11.3 & 2.7 & 37.9 & 10.9 & 8.3 & 7.7 & 4.1 & 42.7 & 21.0 & 18.4 & 17.1 \\
Qwen3-VL-8B-Instruct & 3.3 & 34.0 & 13.2 & 9.3 & 9.3 & 2.0 & 30.2 & 8.7 & 6.0 & 6.1 & 5.4 & 40.0 & 20.2 & 14.6 & 14.2 \\
InternVL3.5-38B & 2.6 & 35.6 & 12.5 & 7.8 & 8.2 & 1.4 & 33.5 & 9.5 & 4.3 & 4.4 & 4.5 & 39.0 & 17.3 & 13.4 & 14.2 \\
InternVL3.5-30B-A3B & 1.1 & 26.4 & 8.8 & 5.6 & 6.2 & 0.7 & 24.6 & 6.6 & 3.0 & 3.8 & 1.9 & 29.3 & 12.2 & 9.7 & 10.1 \\
InternVL3.5-20B-A4B & 1.1 & 22.0 & 8.5 & 5.4 & 5.3 & 1.2 & 20.1 & 6.8 & 2.9 & 2.9 & 1.0 & 25.2 & 11.3 & 9.5 & 9.1 \\
InternVL3.5-14B & 1.1 & 26.0 & 9.3 & 6.0 & 6.1 & 1.2 & 25.0 & 7.0 & 4.0 & 3.8 & 1.0 & 27.6 & 13.0 & 9.1 & 9.9 \\
InternVL3.5-8B & 1.0 & 21.1 & 8.3 & 5.2 & 4.9 & 0.7 & 19.3 & 6.1 & 2.9 & 2.5 & 1.6 & 23.9 & 11.8 & 8.9 & 8.9 \\
LLaVA-OV-8B-RL & 0.4 & 18.8 & 9.0 & 4.2 & 3.2 & 0.1 & 14.1 & 6.5 & 2.1 & 2.0 & 0.8 & 26.2 & 13.0 & 7.4 & 5.2 \\
\bottomrule
\end{tabular*}
\caption{Closed-book VIKR performance (\%), scored under the dual-judge intersection (Gemini-3.1-Pro $\cap$ GPT-5.1). \textsc{Succ.} requires $V{\wedge}I{\wedge}K{\wedge}R$; V/I/K/R are layer-wise coverage. $n{=}1{,}253$ overall, 768 Chinese, 485 English; unscored responses count as failures. Per-judge leaderboards are in Appendix~\ref{app:judge_robustness}.}
\label{tab:main_results}
\end{table*}

\section{Experiments}
\label{sec:experiments}

\subsection{Setup}

We evaluate MemeBench along two complementary axes.
Closed-book evaluation measures image-only interpretation under the standard prompt, while paired augmentation settings test how explanations for the same items change when reasoning guidance or external evidence is added.
All experiments use the full benchmark ($n{=}1{,}253$).

\paragraph{Metrics.}
The primary metric is \textsc{Success} (Eq.~\ref{eq:success}), which requires all four VIKR dimensions to pass.
We also report per-dimension coverage as defined in \S\ref{sec:eval}.

\paragraph{Inference settings.}
The four settings are ordered so that each adds exactly one variable to the previous one.
\emph{Vanilla} gives only the meme image and the standard prompt; \emph{CoT} adds step-by-step reasoning instructions without external evidence.
\emph{Search-CoT} adds generic web evidence through KAR's extractor, search interface, evidence formatting, and grounded reasoner while omitting CultureBase.
\emph{KAR} adds CultureBase entity grounding before web search (\S\ref{sec:kar_method}).
Both retrieval settings use Tavily under the same query budget; adjacent settings thus isolate reasoning guidance, external evidence, and entity grounding in turn.

\paragraph{Models.}
We evaluate 26 models, including nine commercial systems~\citep{google2026gemini31,openai2025gpt5} and open-source families spanning Qwen3-VL~\citep{qwen2025qwen3vl}, InternVL3.5~\citep{wang2025internvl35}, Kimi-K2.5~\citep{moonshot2026kimik25}, MiMo~\citep{xiaomi2025mimovl}, and LLaVA-OneVision~\citep{li2025llavaonevision15}.
This main leaderboard compares general-purpose LVLMs under one image-to-free-form protocol; judge robustness and aggregation sensitivity are reported in Appendix~\ref{app:judge_robustness}.
Table~\ref{tab:main_results} lists them in full.
Figure~\ref{fig:ratio_scaling} summarizes augmentation outcomes for all 26 models; Table~\ref{tab:method_comparison} gives two complementary four-model comparisons.
Gemini-3.1-Pro and Qwen3-VL-235B-A22B-Instruct span the leaderboard's capability range.
GPT-5.6-Sol and Kimi-K2.5 form a matched commercial--open-source pair whose Vanilla profiles agree within 2.0\% on every VIKR dimension, holding the initial diagnosis nearly fixed while model provenance varies.

\begin{table}[b!]
\centering
\scriptsize
\setlength{\tabcolsep}{0pt}
\setlength{\aboverulesep}{0.25ex}
\setlength{\belowrulesep}{0.25ex}
\renewcommand{\arraystretch}{0.94}
\captionsetup{skip=3pt}
\begin{tabular*}{\columnwidth}{@{\extracolsep{\fill}}lcccc@{\hspace{2.5pt}\vrule\hspace{2.5pt}}ccc@{\hspace{2.5pt}\vrule\hspace{2.5pt}}c@{}}
\toprule
\textbf{Model\,/\,Setting} & \textbf{V} & \textbf{I} & \textbf{K} & \textbf{R} & \textsc{Succ.} & \textbf{Rep.} & \textbf{Dmg.} & \textbf{ratio} \\
\midrule
\emph{Gemini-3.1-Pro} & 92.3 & 74.6 & 69.7 & 76.7 & 60.3 & -- & -- & -- \\
\;+ CoT & $-0.1$ & $+1.0$ & $+0.7$ & $+0.8$ & 60.4 & 16.3 & 10.5 & 1.55 \\
\;+ Search-CoT & $-2.9$ & $+3.4$ & $+2.4$ & $+2.2$ & 60.6 & 21.7 & 13.8 & 1.57 \\
\;+ KAR & $-2.5$ & $+5.7$ & $+4.1$ & $+3.7$ & 63.9 & 26.5 & 11.5 & \textbf{2.31} \\
\midrule
\emph{GPT-5.6-Sol} & 72.5 & 58.7 & 55.1 & 60.7 & 40.9 & -- & -- & -- \\
\;+ CoT & $-1.0$ & $+0.5$ & $+0.3$ & $+0.4$ & 40.7 & 11.4 & 16.8 & 0.67 \\
\;+ Search-CoT & $-1.6$ & $+3.9$ & $+3.0$ & $+2.5$ & 44.1 & 17.6 & 17.7 & 0.99 \\
\;+ KAR & $-0.4$ & $+6.4$ & $+5.0$ & $+4.2$ & 47.8 & 21.2 & 13.8 & \textbf{1.54} \\
\midrule
\emph{Kimi-K2.5} & 71.1 & 59.2 & 54.5 & 58.7 & 37.9 & -- & -- & -- \\
\;+ CoT & $-1.1$ & $+0.4$ & $+0.3$ & $+0.3$ & 37.8 & 10.7 & 17.7 & 0.60 \\
\;+ Search-CoT & $-1.5$ & $+3.6$ & $+3.0$ & $+2.3$ & 41.6 & 17.1 & 18.1 & 0.94 \\
\;+ KAR & $-0.2$ & $+6.0$ & $+4.9$ & $+3.9$ & 45.3 & 20.6 & 14.1 & \textbf{1.46} \\
\midrule
\emph{Qwen3-VL-235B-A22B} & 53.2 & 30.6 & 26.1 & 27.3 & 14.5 & -- & -- & -- \\
\;+ CoT & $-2.9$ & $-0.5$ & $-0.4$ & $-0.4$ & 11.4 & 1.4 & 29.7 & 0.05 \\
\;+ Search-CoT & $-6.8$ & $+1.2$ & $+2.8$ & $+0.8$ & 17.7 & 8.0 & 25.3 & 0.32 \\
\;+ KAR & $-9.0$ & $+2.1$ & $+4.7$ & $+1.4$ & 19.7 & 9.2 & 18.1 & \textbf{0.50} \\
\bottomrule
\end{tabular*}
\caption{Controlled augmentation on four representative models, showing component-level changes and preservation of Vanilla performance. Baseline rows give absolute VIKR and \textsc{Success}; augmented rows give VIKR changes and absolute \textsc{Success}. All entries except \textbf{ratio} are in \%. \textbf{Rep.}/\textbf{Dmg.} are shares of Vanilla failures repaired/passes broken; \textbf{ratio} is their quotient. Qwen3-VL-235B-A22B is Instruct.}
\label{tab:method_comparison}
\end{table}

\subsection{Closed-Book Results}
\label{sec:main_results}
\label{sec:baseline}

Table~\ref{tab:main_results} reports performance under a common image-only, closed-book protocol.
On this measure, MemeBench remains far from saturated.
Gemini-3.1-Pro leads with 60.3\% \textsc{Success}, followed by Gemini-3-Flash (53.7\%) and GPT-5.6-Sol (40.9\%).
Claude-Opus-5 follows at 38.3\%, while Kimi-K2.5 is the strongest open-source model at 37.9\%, only 3.0\% behind GPT-5.6-Sol; every model outside the top three remains below 40\%.
Within the GPT-5.6 family, Sol (40.9\%), Terra (27.8\%), and Luna (18.0\%) are consistently ordered across \textsc{Success} and all four dimensions, with the largest separations appearing in Identity, Knowledge, and Reasoning.
Within Qwen3-VL, Instruct leads at 235B and 32B, while Thinking leads at 30B and ties at 8B; Thinking therefore does not automatically yield higher complete VIKR coverage.
Closed-book performance also differs across the two sampled ecosystems.
Gemini-3.1-Pro scores 76.7\% on English and 49.9\% on Chinese, while GPT-5.6-Terra scores 36.5\% and 22.3\%, respectively.
The same direction holds for GPT-5.6-Sol and Kimi-K2.5, showing that the aggregate difference is not confined to one model family.
The six models above $65\%$ Visual coverage still span $28.4$--$60.3\%$ \textsc{Success}, motivating the layer-wise diagnosis below.

\subsection{VIKR Diagnostics}
\label{sec:diagnostic}

\begin{figure}[t!]
  \centering
  \includegraphics[width=0.95\columnwidth]{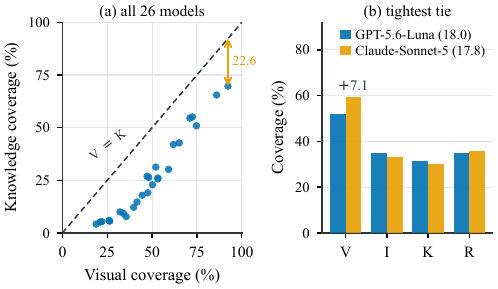}
  \caption{\textbf{VIKR diagnostics.} \textbf{(a)} Visual--Knowledge coverage across 26 models; the arrow marks the strongest model's 22.6\% gap. \textbf{(b)} Closest \textsc{Success} pair, 0.2\% apart.}
  \label{fig:vikr_diagnostics}
\end{figure}

Every model sits on the same side of the Visual-Knowledge diagonal (Figure~\ref{fig:vikr_diagnostics}(a)).
The gap spans 14.6 to 29.0\% across the evaluated models and remains 22.6\% for the strongest, showing that Knowledge remains a bottleneck across capability tiers.
Layer-wise scoring also separates models a single score ranks together.
The tightest \textsc{Success} tie in Table~\ref{tab:main_results} is GPT-5.6-Luna and Claude-Sonnet-5, 0.2\% apart, yet their failures sit in different places (Figure~\ref{fig:vikr_diagnostics}(b)).
Claude-Sonnet-5 covers 7.1\% more visual content while grounding 1.6\% fewer entities and 1.1\% less background knowledge, so its bottleneck is cultural grounding, not perception.
The same diagnostic separation appears near the top of the leaderboard: Claude-Opus-5 and Kimi-K2.5 are only 0.4\% apart in \textsc{Success}, but Opus covers 3.6\% more Visual content and 8.8\% fewer Identity links, and their Visual--Knowledge gaps differ by 7.2\%.
Across Table~\ref{tab:main_results}, seven model pairs lie within 0.6\% in \textsc{Success} while their Visual--Knowledge gaps differ by at least 4\%.

The same resolution appears within individual models.
Responses that identify an entity without supplying the required background ($I{=}1,K{=}0$) account for a median 13.0\% of responses.
The reverse pattern ($I{=}0,K{=}1$) accounts for 5.2\%, while responses that satisfy both but miss the interpretation ($I{=}1,K{=}1,R{=}0$) account for 2.2\%.
Across the 26 Vanilla runs, the pooled responses realize all 16 VIKR states, including $0001$ and $0010$, with non-monotonic states accounting for 10.9\% to 29.2\% per model (Appendix~\ref{app:states}).
These patterns preserve response-level distinctions that a single \textsc{Success} score collapses.
Identity and Reasoning also track one another closely---within 4\% for most models---while remaining separable in the observed state patterns.

Cross-language results further localize the gap.
From English to Chinese, Gemini-3.1-Pro drops 4.9\% on Visual but 20.2\% on Identity and 22.2\% on Knowledge; for GPT-5.6-Terra, the Knowledge gap is 22.7\% versus 12.3\% on Visual.
Thus, the larger differences concentrate in the culturally loaded dimensions.
Because the subsets differ in platform and domain composition, including ACG shares of 59\% in Chinese and 36\% in English, they characterize two sampled online-cultural ecosystems rather than a causal language effect.

This diagnostic structure yields a testable prediction.
If VIKR captures distinguishable response requirements, targeted evidence should selectively change the diagnosed layers rather than move all four dimensions together.

\FloatBarrier

\subsection{Retrieval Analysis}
\label{sec:retrieval_mitigation}

Across the four models, KAR raises Knowledge by $4.1$ to $5.0$\% and Identity by up to $6.4$\%, directly moving the diagnosed bottlenecks.
CoT changes \textsc{Success} by at most $3.1$\% without a consistent direction, whereas both retrieval settings improve Identity and Knowledge.

\begin{figure}[t!]
\centering
\includegraphics[width=0.95\columnwidth]{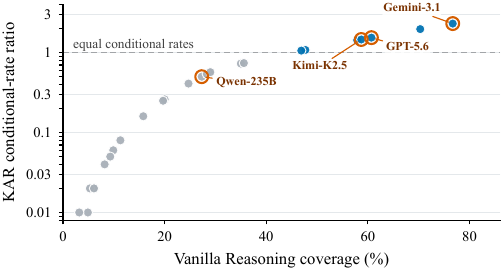}
\caption{\textbf{KAR repair--damage balance.}
Ratio of the repair rate among Vanilla failures to the damage rate among Vanilla passes under KAR, versus Vanilla Reasoning coverage across 26 models.
Blue/gray points fall above/below $1$; the dashed line marks equal rates, and labels mark the four detailed models.}
\label{fig:ratio_scaling}
\end{figure}

Paired transitions expose what net \textsc{Success} conceals.
Under KAR, Gemini-3.1-Pro repairs $26.5\%$ of its failures and breaks $11.5\%$ of its passes; Qwen3-VL-235B-A22B-Instruct repairs $9.2\%$ and breaks $18.1\%$.
Yet Qwen gains more net \textsc{Success} ($+5.2\%$ versus $+3.6\%$), reversing the ranking by preservation quality.
Across 26 models, ratio estimates are non-decreasing from CoT to Search-CoT to KAR, with 21 strict increases.
Under KAR, the seven estimates above $1$ are exactly the seven models with highest Vanilla Reasoning coverage (Figure~\ref{fig:ratio_scaling}).
Because the rates condition on different Vanilla subsets, the ratio measures preservation balance, not net efficacy; every KAR $\Delta$\textsc{Success} estimate remains positive.
The comparison therefore emphasizes preservation quality rather than only average score changes.

The layer transitions explain why these rankings differ.
Across all eight retrieval comparisons, Identity and Knowledge rise while Visual falls; Reasoning also rises, but by a smaller margin.
These opposing movements provide direct response-level evidence that VIKR captures distinct changes that a single aggregate score cannot expose.

Because Search-CoT matches KAR except for CultureBase, their comparison tests entity grounding under controlled retrieval conditions.
KAR repairs more items and breaks fewer than generic web search on every detailed model.
For Gemini-3.1-Pro, Search-CoT leaves the ratio nearly unchanged ($1.55$ to $1.57$), whereas adding CultureBase raises it to $2.31$, showing the value of entity grounding in this controlled comparison.
Appendix~\ref{app:culturebase} details CultureBase construction and the entity-grounding stage.
The matched GPT-5.6-Sol--Kimi-K2.5 pair tests augmentation from near-identical initial VIKR profiles across model provenance; their final ratios differ by only $0.08$.

Inspection of pass-to-fail cases suggests \emph{evidence crowding}, where retrieved context displaces image-grounded or task-relevant content.
Qwen3-VL-235B-A22B-Instruct provides the clearest example, losing $9.0$\% of Visual coverage under KAR.
Thus, MemeBench tests whether evidence fills diagnosed gaps without displacing coverage the model already had.

\section{Conclusion}
\label{sec:conclusion}

We introduced MemeBench, a bilingual diagnostic benchmark of 1,253 Chinese and English memes centered on ACG and adjacent online cultures.
VIKR evaluates open-ended explanations through Visual, Identity, Knowledge, and Reasoning requirements, reserving \textsc{Success} for complete interpretations.
Across 26 LVLMs, every model covers visible content more reliably than required knowledge; even the strongest retains a 22.6\% Visual--Knowledge gap.
Layer-wise profiles also distinguish models with similar aggregate scores.
Paired augmentation shows why this diagnosis matters: retrieval improves Identity and Knowledge but can reduce Visual coverage, while KAR achieves a better repair--damage balance than generic retrieval across all four models.
MemeBench identifies missing content and tests whether evidence fills it, extending evaluation beyond a single score.

\bibliography{references}

@article{kiela2020hateful,
  title={The hateful memes challenge: Detecting hate speech in multimodal memes},
  author={Kiela, Douwe and Firooz, Hamed and Mohan, Aravind and Goswami, Vedanuj and Singh, Amanpreet and Ringshia, Pratik and Testuggine, Davide},
  journal={Proc. of NeurIPS},
  volume={33},
  pages={2611--2624},
  year={2020}
}

@inproceedings{hwang2023memecap,
  title={Memecap: A dataset for captioning and interpreting memes},
  author={Hwang, EunJeong and Shwartz, Vered},
  booktitle={Proc. of EMNLP},
  pages={1433--1445},
  year={2023}
}

@article{lewis2020rag,
  title={Retrieval-augmented generation for knowledge-intensive nlp tasks},
  author={Lewis, Patrick and Perez, Ethan and Piktus, Aleksandra and Petroni, Fabio and Karpukhin, Vladimir and Goyal, Naman and K{\"u}ttler, Heinrich and Lewis, Mike and Yih, Wen-tau and Rockt{\"a}schel, Tim and others},
  journal={Proc. of NeurIPS},
  volume={33},
  pages={9459--9474},
  year={2020}
}

@article{chen2024bge,
  title={{BGE M3-Embedding}: Multi-Lingual, Multi-Functionality, Multi-Granularity Text Embeddings through Self-Knowledge Distillation},
  author={Chen, Jianlv and Xiao, Shitao and Zhang, Peitian and Luo, Kun and Lian, Defu and Liu, Zheng},
  journal={arXiv:2402.03216},
  volume={4},
  number={5},
  year={2024}
}

@inproceedings{pramanick2021momenta,
  title={MOMENTA: A multimodal framework for detecting harmful memes and their targets},
  author={Pramanick, Shraman and Sharma, Shivam and Dimitrov, Dimitar and Akhtar, Md Shad and Nakov, Preslav and Chakraborty, Tanmoy},
  booktitle={Findings of EMNLP},
  pages={4439--4455},
  year={2021}
}

@inproceedings{marino2019okvqa,
  title={Ok-vqa: A visual question answering benchmark requiring external knowledge},
  author={Marino, Kenneth and Rastegari, Mohammad and Farhadi, Ali and Mottaghi, Roozbeh},
  booktitle={Proc. of CVPR},
  pages={3195--3204},
  year={2019}
}

@article{openai2025gpt5,
  title={{OpenAI} {GPT}-5 System Card},
  author={Singh, Aaditya and Fry, Adam and Perelman, Adam and Tart, Adam and Ganesh, Adi and El-Kishky, Ahmed and McLaughlin, Aidan and Low, Aiden and Ostrow, AJ and Ananthram, Akhila and others},
  journal={arXiv:2601.03267},
  year={2026}
}

@techreport{google2026gemini31,
  title={Gemini 3.1 Pro Model Card},
  author={{Google DeepMind}},
  year={2026},
  institution={Google DeepMind},
  note={\url{https://deepmind.google/models/model-cards/gemini-3-1-pro/}}
}

@article{qwen2025qwen3vl,
  title={{Qwen3-VL} Technical Report},
  author={Bai, Shuai and Cai, Yuxuan and Chen, Ruizhe and Chen, Keqin and Chen, Xionghui and Cheng, Zesen and Deng, Lianghao and Ding, Wei and Gao, Chang and Ge, Chunjiang and others},
  journal={arXiv:2511.21631},
  year={2025}
}

@inproceedings{kasu2025dhumor,
  title={D-humor: Dark humor understanding via multimodal open-ended reasoning},
  author={Kasu, Sai Kartheek Reddy and Rehman, Mohammad Zia Ur and Dar, Shahid Shafi and Junghare, Rishi Bharat and Namboodiri, Dhanvin Sanjay and Kumar, Nagendra},
  booktitle={Proc. of ICDM},
  pages={377--386},
  year={2025},
  organization={IEEE}
}

@article{degiorgis2026mquest,
  title={M-QUEST--Meme Question-Understanding Evaluation on Semantics and Toxicity},
  author={De Giorgis, Stefano and Chen, Ting-Chih and Ilievski, Filip},
  journal={arXiv:2603.03315},
  year={2026}
}

@inproceedings{chen2022murag,
  title={Murag: Multimodal retrieval-augmented generator for open question answering over images and text},
  author={Chen, Wenhu and Hu, Hexiang and Chen, Xi and Verga, Pat and Cohen, William},
  booktitle={Proc. of EMNLP},
  pages={5558--5570},
  year={2022}
}

@inproceedings{caffagni2024wikillava,
  title={Wiki-llava: Hierarchical retrieval-augmented generation for multimodal llms},
  author={Caffagni, Davide and Cocchi, Federico and Moratelli, Nicholas and Sarto, Sara and Cornia, Marcella and Baraldi, Lorenzo and Cucchiara, Rita},
  booktitle={Proc. of CVPR},
  pages={1818--1826},
  year={2024}
}

@article{moonshot2026kimik25,
  title={{Kimi K2.5}: Visual Agentic Intelligence},
  author={Team, Kimi and Bai, Tongtong and Bai, Yifan and Bao, Yiping and Cai, SH and Cao, Yuan and Charles, Y and Che, HS and Chen, Cheng and Chen, Guanduo and others},
  journal={arXiv:2602.02276},
  year={2026}
}

@article{wang2025internvl35,
  title={{InternVL3.5}: Advancing Open-Source Multimodal Models in Versatility, Reasoning, and Efficiency},
  author={Wang, Weiyun and Gao, Zhangwei and Gu, Lixin and Pu, Hengjun and Cui, Long and Wei, Xingguang and Liu, Zhaoyang and Jing, Linglin and Ye, Shenglong and Shao, Jie and others},
  journal={arXiv:2508.18265},
  year={2025}
}

@misc{xiaomi2025mimovl,
      title={MiMo-VL Technical Report}, 
      author={Core Team and Zihao Yue and Zhenru Lin and Yifan Song and Weikun Wang and Shuhuai Ren and Shuhao Gu and Shicheng Li and Peidian Li and Liang Zhao and Lei Li and Kainan Bao and Hao Tian and Hailin Zhang and Gang Wang and Dawei Zhu and Cici and Chenhong He and Bowen Ye and Bowen Shen and Zihan Zhang and Zihan Jiang and Zhixian Zheng and Zhichao Song and Zhenbo Luo and Yue Yu and Yudong Wang and Yuanyuan Tian and Yu Tu and Yihan Yan and Yi Huang and Xu Wang and Xinzhe Xu and Xingchen Song and Xing Zhang and Xing Yong and Xin Zhang and Xiangwei Deng and Wenyu Yang and Wenhan Ma and Weiwei Lv and Weiji Zhuang and Wei Liu and Sirui Deng and Shuo Liu and Shimao Chen and Shihua Yu and Shaohui Liu and Shande Wang and Rui Ma and Qiantong Wang and Peng Wang and Nuo Chen and Menghang Zhu and Kangyang Zhou and Kang Zhou and Kai Fang and Jun Shi and Jinhao Dong and Jiebao Xiao and Jiaming Xu and Huaqiu Liu and Hongshen Xu and Heng Qu and Haochen Zhao and Hanglong Lv and Guoan Wang and Duo Zhang and Dong Zhang and Di Zhang and Chong Ma and Chang Liu and Can Cai and Bingquan Xia},
      year={2025},
      eprint={2506.03569},
      archivePrefix={arXiv},
      primaryClass={cs.CL},
      url={https://arxiv.org/abs/2506.03569}, 
}

@article{li2025llavaonevision15,
  title={Llava-onevision-1.5: Fully open framework for democratized multimodal training},
  author={An, Xiang and Xie, Yin and Yang, Kaicheng and Zhang, Wenkang and Zhao, Xiuwei and Cheng, Zheng and Wang, Yirui and Xu, Songcen and Chen, Changrui and Zhu, Didi and others},
  journal={arXiv:2509.23661},
  year={2025}
}

@article{zheng2023judging,
  title={Judging llm-as-a-judge with mt-bench and chatbot arena},
  author={Zheng, Lianmin and Chiang, Wei-Lin and Sheng, Ying and Zhuang, Siyuan and Wu, Zhanghao and Zhuang, Yonghao and Lin, Zi and Li, Zhuohan and Li, Dacheng and Xing, Eric and others},
  journal={Proc. of NeurIPS},
  volume={36},
  pages={46595--46623},
  year={2023}
}

@inproceedings{chen2023infoseek,
  title={Can pre-trained vision and language models answer visual information-seeking questions?},
  author={Chen, Yang and Hu, Hexiang and Luan, Yi and Sun, Haitian and Changpinyo, Soravit and Ritter, Alan and Chang, Ming-Wei},
  booktitle={Proc. of EMNLP},
  pages={14948--14968},
  year={2023}
}

@inproceedings{mensink2023encyclopedic,
  title={Encyclopedic vqa: Visual questions about detailed properties of fine-grained categories},
  author={Mensink, Thomas and Uijlings, Jasper and Castrejon, Lluis and Goel, Arushi and Cadar, Felipe and Zhou, Howard and Sha, Fei and Araujo, Andr{\'e} and Ferrari, Vittorio},
  booktitle={Proc. of ICCV},
  pages={3113--3124},
  year={2023}
}

@inproceedings{cohen2025entitygap,
  title={Performance gap in entity knowledge extraction across modalities in vision language models},
  author={Cohen, Ido and Gottesman, Daniela and Geva, Mor and Giryes, Raja},
  booktitle={Proc. of ACL},
  pages={29095--29108},
  year={2025}
}

@article{zhou2026figurative,
  title={I Came, I Saw, I Explained: Benchmarking Multimodal LLMs on Figurative Meaning in Memes},
  author={Zhou, Shijia and Mohammad, Saif M and Plank, Barbara and Frassinelli, Diego},
  journal={arXiv:2603.23229},
  year={2026}
}

@inproceedings{bui2024multi3hate,
  title={Multi3Hate: Multimodal, multilingual, and multicultural hate speech detection with vision--language models},
  author={Bui, Minh Duc and Von Der Wense, Katharina and Lauscher, Anne},
  booktitle={Proceedings of the 2025 Conference of the Nations of the Americas Chapter of the Association for Computational Linguistics: Human Language Technologies (Volume 1: Long Papers)},
  pages={9714--9731},
  year={2025}
}

@article{yadav2025cultural,
  title={Evaluation of Cultural Competence of Vision-Language Models},
  author={Yadav, Srishti and Tilton, Lauren and Antoniak, Maria and Arnold, Taylor and Li, Jiaang and Pawar, Siddhesh Milind and Karamolegkou, Antonia and Frank, Stella and An, Zhaochong and Rostamzadeh, Negar and others},
  journal={arXiv:2505.22793},
  year={2025}
}

@inproceedings{zhao2025memereacon,
  title={MemeReaCon: Probing Contextual Meme Understanding in Large Vision-Language Models},
  author={Zhao, Zhengyi and Zhang, Shubo and Zhang, Yuxi and Zhao, Yanxi and Zhang, Yifan and Wang, Zezhong and Wang, Huimin and Zhao, Yutian and Liang, Bin and Zheng, Yefeng and others},
  booktitle={Proc. of EMNLP},
  pages={3559--3582},
  year={2025}
}

@article{zhao2026transcreation,
  title={Beyond Translation: Cross-Cultural Meme Transcreation with Vision-Language Models},
  author={Zhao, Yuming and Zhang, Peiyi and Ignat, Oana},
  journal={arXiv:2602.02510},
  year={2026}
}

@inproceedings{liu2024mmbench,
  title={Mmbench: Is your multi-modal model an all-around player?},
  author={Liu, Yuan and Duan, Haodong and Zhang, Yuanhan and Li, Bo and Zhang, Songyang and Zhao, Wangbo and Yuan, Yike and Wang, Jiaqi and He, Conghui and Liu, Ziwei and others},
  booktitle={Proc. of ECCV},
  pages={216--233},
  year={2024},
  organization={Springer}
}

@inproceedings{yue2024mmmu,
  title={Mmmu: A massive multi-discipline multimodal understanding and reasoning benchmark for expert agi},
  author={Yue, Xiang and Ni, Yuansheng and Zhang, Kai and Zheng, Tianyu and Liu, Ruoqi and Zhang, Ge and Stevens, Samuel and Jiang, Dongfu and Ren, Weiming and Sun, Yuxuan and others},
  booktitle={Proc. of CVPR},
  pages={9556--9567},
  year={2024}
}

@article{yu2024mmvet,
  title={Mm-vet: Evaluating large multimodal models for integrated capabilities},
  author={Yu, Weihao and Yang, Zhengyuan and Li, Linjie and Wang, Jianfeng and Lin, Kevin and Liu, Zicheng and Wang, Xinchao and Wang, Lijuan},
  journal={arXiv:2308.02490},
  year={2023}
}

@article{fu2023mme,
  title={Mme: A comprehensive evaluation benchmark for multimodal large language models},
  author={Fu, Chaoyou and Chen, Peixian and Shen, Yunhang and Qin, Yulei and Zhang, Mengdan and Lin, Xu and Yang, Jinrui and Zheng, Xiawu and Li, Ke and Sun, Xing and others},
  journal={arXiv:2306.13394},
  year={2023}
}

@article{lu2024toxicnmm,
  title={Towards comprehensive detection of chinese harmful memes},
  author={Lu, Junyu and Xu, Bo and Zhang, Xiaokun and Wang, Hongbo and Zhu, Haohao and Zhang, Dongyu and Yang, Liang and Lin, Hongfei},
  journal={Proc. of NeurIPS},
  volume={37},
  pages={13302--13320},
  year={2024}
}

@inproceedings{nguyen2025memeqa,
  title={MemeQA: Holistic evaluation for meme understanding},
  author={Nguyen, Khoi PN and Li, Terrence and Zhou, Derek Lou and Xiong, Gabriel and Balu, Pranav and Alahari, Nandhan and Huang, Alan and Chauhan, Tanush and Bala, Harshavardhan and Guzelordu, Emre and others},
  booktitle={Proc. of ACL},
  pages={18926--18946},
  year={2025}
}

@inproceedings{chiu2025culturalbench,
  title={CulturalBench: A robust, diverse and challenging benchmark for measuring LMs’ cultural knowledge through human-AI red-teaming},
  author={Chiu, Yu Ying and Jiang, Liwei and Lin, Bill Yuchen and Park, Chan Young and Li, Shuyue Stella and Ravi, Sahithya and Bhatia, Mehar and Antoniak, Maria and Tsvetkov, Yulia and Shwartz, Vered and others},
  booktitle={Proc. of ACL},
  pages={25663--25701},
  year={2025}
}

@inproceedings{zhang2025ciibench,
  title={Can MLLMs Understand the Deep Implication Behind Chinese Images?},
  author={Zhang, Chenhao and Feng, Xi and Bai, Yuelin and Du, Xeron and Hou, Jinchang and Deng, Kaixin and Han, Guangzeng and Li, Qinrui and Wang, Bingli and Liu, Jiaheng and others},
  booktitle={Proc. of ACL},
  pages={14369--14402},
  year={2025}
}

@inproceedings{lee2025checkeval,
  title={Checkeval: A reliable llm-as-a-judge framework for evaluating text generation using checklists},
  author={Lee, Yukyung and Kim, Joonghoon and Kim, Jaehee and Cho, Hyowon and Kang, Jaewook and Kang, Pilsung and Kim, Najoung},
  booktitle={Proc. of EMNLP},
  pages={15782--15809},
  year={2025}
}

@inproceedings{hu2023oven,
  title={Open-domain visual entity recognition: Towards recognizing millions of wikipedia entities},
  author={Hu, Hexiang and Luan, Yi and Chen, Yang and Khandelwal, Urvashi and Joshi, Mandar and Lee, Kenton and Toutanova, Kristina and Chang, Ming-Wei},
  booktitle={Proc. of ICCV},
  pages={12065--12075},
  year={2023}
}

@article{li2024searchlvlms,
  title={Searchlvlms: A plug-and-play framework for augmenting large vision-language models by searching up-to-date internet knowledge},
  author={Li, Chuanhao and Li, Zhen and Jing, Chenchen and Liu, Shuo and Shao, Wenqi and Wu, Yuwei and Luo, Ping and Qiao, Yu and Zhang, Kaipeng},
  journal={Proc. of NeurIPS},
  volume={37},
  pages={64582--64603},
  year={2024}
}

@inproceedings{nandy2024yesbut,
  title={*** YesBut***: A High-Quality Annotated Multimodal Dataset for evaluating Satire Comprehension capability of Vision-Language Models},
  author={Nandy, Abhilash and Agarwal, Yash and Patwa, Ashish and Das, Millon Madhur and Bansal, Aman and Raj, Ankit and Goyal, Pawan and Ganguly, Niloy},
  booktitle={Proc. of EMNLP},
  pages={16878--16895},
  year={2024}
}

@article{wu2025mmsearchr1,
  title={Mmsearch-r1: Incentivizing lmms to search},
  author={Wu, Jinming and Deng, Zihao and Li, Wei and Liu, Yiding and You, Bo and Li, Bo and Ma, Zejun and Liu, Ziwei},
  journal={arXiv:2506.20670},
  year={2025}
}

@article{zhang2026vsearcher,
  title={VSearcher: Long-Horizon Multimodal Search Agent via Reinforcement Learning},
  author={Zhang, Ruiyang and Sun, Qianguo and Song, Chao and Qi, Yiyan and Zheng, Zhedong},
  journal={arXiv:2603.02795},
  year={2026}
}

@article{myung2024blend,
  title={Blend: A benchmark for llms on everyday knowledge in diverse cultures and languages},
  author={Myung, Junho and Lee, Nayeon and Zhou, Yi and Jin, Jiho and Putri, Rifki A and Antypas, Dimosthenis and Borkakoty, Hsuvas and Kim, Eunsu and Perez-Almendros, Carla and Ayele, Abinew A and others},
  journal={Proc. of NeurIPS},
  volume={37},
  pages={78104--78146},
  year={2024}
}

@inproceedings{park2024memeintent,
  title={MemeIntent: Benchmarking intent description generation for memes},
  author={Park, Jeongsik and Nguyen, Khoi PN and Li, Terrence and Shrestha, Suyesh and Vu, Megan Kim and Wang, Jerry Yining and Ng, Vincent},
  booktitle={Proceedings of the 25th Annual Meeting of the Special Interest Group on Discourse and Dialogue},
  pages={631--643},
  year={2024}
}

@inproceedings{xu2026metagpt,
  title={MetaGPT: A Large Vision-Language Model for Meme Metaphor Understanding},
  author={Xu, Bo and Wang, Chenyuan and Chen, Xinyu and Lin, Hongfei and Xia, Feng},
  booktitle={Proceedings of the AAAI Conference on Artificial Intelligence},
  volume={40},
  number={19},
  pages={16040--16048},
  year={2026}
}

@article{shahroor2026memelens,
  title={MemeLens: Multilingual Multitask VLMs for Memes},
  author={Shahroor, Ali Ezzat and Kmainasi, Mohamed Bayan and Hasnat, Abul and Dimitrov, Dimitar and Martino, Giovanni Da San and Nakov, Preslav and Alam, Firoj},
  journal={arXiv preprint arXiv:2601.12539},
  year={2026}
}

@inproceedings{wang2026nativememes,
  title={From Native Memes to Global Moderation: Cross-Cultural Evaluation of Vision--Language Models for Hateful Meme Detection},
  author={Wang, Mo and Ren, Kaixuan and Jalan, Pratik and Ashraf, Ahmed and Vu, Tuong Vy and Seetharaman, Rahul and Nawaz, Shah and Naseem, Usman},
  booktitle={Proceedings of the ACM Web Conference 2026},
  pages={9788--9799},
  year={2026}
}

@inproceedings{nayak2024culturalvqa,
  title={Benchmarking vision language models for cultural understanding},
  author={Nayak, Shravan and Jain, Kanishk and Awal, Rabiul and Reddy, Siva and Van Steenkiste, Sjoerd and Hendricks, Lisa Anne and Sta{\'n}czak, Karolina and Agrawal, Aishwarya},
  booktitle={Proceedings of the 2024 Conference on Empirical Methods in Natural Language Processing},
  pages={5769--5790},
  year={2024}
}

@inproceedings{tan2026blendvis,
  title={Blend-vis: Benchmarking multimodal cultural understanding in vision language models},
  author={Tan, Bryan Chen Zhengyu and Zheng, Weihua and Liu, Zhengyuan and Chen, Nancy and Lee, Hwaran and Choo, Kenny Tsu Wei and Lee, Roy Ka-Wei},
  booktitle={Proceedings of the 19th Conference of the European Chapter of the Association for Computational Linguistics (Volume 1: Long Papers)},
  pages={4647--4669},
  year={2026}
}

@article{jiang2026avmeme,
  title={AVMeme Exam: A Multimodal Multilingual Multicultural Benchmark for LLMs' Contextual and Cultural Knowledge and Thinking},
  author={Jiang, Xilin and Wang, Qiaolin and Wu, Junkai and He, Xiaomin and Xu, Zhongweiyang and Ma, Yinghao and Piao, Minshuo and Yang, Kaiyi and Zheng, Xiuwen and Shimizu, Riki and others},
  journal={arXiv preprint arXiv:2601.17645},
  year={2026}
}

@inproceedings{zhong2024fime,
  title={Multimodal understanding of memes with fair explanations},
  author={Zhong, Yang and Baghel, Bhiman Kumar},
  booktitle={Proceedings of the IEEE/CVF Conference on Computer Vision and Pattern Recognition},
  pages={2007--2017},
  year={2024}
}

@inproceedings{xu2025punmemecn,
  title={PUNMEMECN: A Benchmark to Explore Vision-Language Models’ Understanding of Chinese Pun Memes},
  author={Xu, Zhijun and Yuan, Siyu and Zhang, Yiqiao and Sun, Jingyu and Zheng, Tong and Yang, Deqing},
  booktitle={Proceedings of the 2025 Conference on Empirical Methods in Natural Language Processing},
  pages={18705--18721},
  year={2025}
}

\clearpage
\appendix
\setcounter{secnumdepth}{2}

\raggedbottom
\FloatBarrier

\section{Benchmark Construction and Human Validation}
\label{app:dataquality}

\subsection{Dataset Selection and Composition}
\label{app:selection}

Quality control removes 247 of the 1,500 candidates (\S\ref{sec:data}).
Table~\ref{tab:selection} shows the domain and language distributions of the retained set.
\begin{table}[t]
\centering
\small
\begin{tabular}{@{}lrrrc@{}}
\toprule
\textbf{Domain} & \textbf{ZH} & \textbf{EN} & \textbf{Total} & \textbf{\%} \\
\midrule
ACG           & 454 & 174 & 628 & 50.1 \\
Cross-Domain  & 142 & 124 & 266 & 21.2 \\
Daily Life    &  91 &  75 & 166 & 13.2 \\
Movies \& TV  &  47 &  78 & 125 & 10.0 \\
Public Life   &  34 &  34 &  68 &  5.4 \\
\midrule
\textbf{Total} & \textbf{768} & \textbf{485} & \textbf{1,253} & \\
\bottomrule
\end{tabular}
\caption{Joint domain-by-language distribution of the retained evaluation set ($n{=}1{,}253$); these are the counts plotted in Figure~\ref{fig:benchmark_stats}.}
\label{tab:selection}
\end{table}

\subsection{Human Study and Calibration}
\label{app:human_study}

\paragraph{Design, roles, and blinding.}
\label{app:human_design}

We use one \HumanSubsetN-item study to validate reference independence and calibrate the LLM judges.
The subset is sampled before annotation review and stratified jointly by language and analysis domain, preserving the benchmark proportions up to integer rounding.
The protocol assigns four participants to role-separated responsibilities:
\begin{enumerate}[leftmargin=*,nosep]
    \item \textbf{P1 (primary annotator)} authored the original explanations and metadata for the full candidate pool.
    \item \textbf{P2 and P3 (blind respondents)} each answer all \HumanSubsetN{} memes using the same explanation prompt as evaluated models. They see only the meme and prompt---not the existing reference, VIKR fields, checklist, model outputs, or one another's answer.
    \item \textbf{P4 (independent validator)} did not create the original references and did not contribute blind explanations. P4 compares the two blind explanations with the existing reference and adjudicates substantive semantic conflicts.
\end{enumerate}
P2 and P3 are bilingual Chinese--English speakers familiar with online meme culture; language proficiency and self-reported ACG familiarity are recorded before the study.
All \HumanResponseN{} blind answers are frozen before any participant is shown a reference or checklist.

\paragraph{Reference validation and adjudication.}
\label{app:reference_validation_protocol}

P4 compares each pair of blind explanations with the existing reference on three content-bearing criteria: core entity identity, required cultural background, and humor or communicative mechanism.
Each item is assigned one of four outcomes:
\emph{fully compatible} (same interpretation and required facts),
\emph{partially compatible} (same core reading with complementary or nonessential differences),
\emph{materially conflicting} (incompatible entity, background, or mechanism), or
\emph{indeterminate} (insufficient evidence to select a reading).
For every material conflict, P4 reviews the image and external cultural sources, records a rationale, and chooses among retaining the existing reference, revising the reference/checklist to admit a supported reading, or removing an item that lacks a defensible evaluation target.
References and checklists are frozen only after these decisions.

\begin{table}[t]
\centering
\small
\begin{tabular}{@{}lrr@{}}
\toprule
\textbf{Compatibility outcome} & \textbf{Count} & \textbf{\%} \\
\midrule
Fully compatible      & \RefFullN      & \RefFullPct \\
Partially compatible  & \RefPartialN   & \RefPartialPct \\
Materially conflicting& \RefConflictN  & \RefConflictPct \\
Indeterminate         & \RefUncertainN & \RefUncertainPct \\
\midrule
Total                 & \HumanSubsetN  & 100.0 \\
\bottomrule
\end{tabular}
\caption{Independent reference validation on the stratified human-study subset. Adjudication results in \RefRevisedN{} revised references and \RefRemovedN{} removed items.}
\label{tab:reference_validation}
\end{table}

\paragraph{Human labels and judge calibration.}
\label{app:judge_calibration}

To calibrate automatic evaluation on model-like outputs, we additionally select one frozen candidate response per study item (\JudgeCalibrationN{} responses total), balanced across model families and performance bands before human labeling.
Model identity and both LLM decisions are hidden.
P1 and P4 independently label every checklist item; their pre-adjudication agreement is reported below, and disagreements are resolved by discussion to produce the human gold labels.
Neither rater evaluates a response they authored.
We then compare Gemini-3.1-Pro, GPT-5.1, and their intersection separately with these gold labels using Cohen's $\kappa$.

\begin{table}[t]
\centering
\small
\setlength{\tabcolsep}{3.5pt}
\begin{tabular}{@{}lccccc@{}}
\toprule
\textbf{Comparison} & \textbf{V} & \textbf{I} & \textbf{K} & \textbf{R} & \textsc{Success} \\
\midrule
Human--human
& \HumanHumanV & \HumanHumanI & \HumanHumanK & \HumanHumanR & \HumanHumanSuccess \\
Gemini--human
& \GeminiHumanV & \GeminiHumanI & \GeminiHumanK & \GeminiHumanR & \GeminiHumanSuccess \\
GPT--human
& \GPTHumanV & \GPTHumanI & \GPTHumanK & \GPTHumanR & \GPTHumanSuccess \\
Intersection--human
& \IntersectionHumanV & \IntersectionHumanI & \IntersectionHumanK & \IntersectionHumanR & \IntersectionHumanSuccess \\
\bottomrule
\end{tabular}
\caption{Pre-adjudication human--human agreement and judge--human agreement (Cohen's $\kappa$) on \JudgeCalibrationN{} frozen model responses.}
\label{tab:judge_human}
\end{table}

\paragraph{Human reference performance.}
\label{app:human_baseline}

Because the blind explanations of \S\ref{app:human_design} answer the same prompt given to the evaluated models, scoring them against the same VIKR checklists under the same conjunctive \textsc{Success} definition places human performance on the benchmark's own scale.
Table~\ref{tab:human_baseline} reports it alongside the strongest model.

\begin{table}[t]
\centering
\small
\begin{tabular}{@{}lccccc@{}}
\toprule
& \textbf{V} & \textbf{I} & \textbf{K} & \textbf{R} & \textsc{Success} \\
\midrule
Human & \HumanBaselineV & \HumanBaselineI & \HumanBaselineK & \HumanBaselineR & \HumanBaselineSuccess \\
Gemini-3.1-Pro & 92.3 & 74.6 & 69.7 & 76.7 & 60.3 \\
\midrule
Gap & $+4.7$ & $+19.4$ & $+20.3$ & $+14.3$ & $+22.7$ \\
\bottomrule
\end{tabular}
\caption{Human reference performance on the study subset (\%), scored on the same checklists as model responses, against the leading model from Table~\ref{tab:judge_robustness}. Human \textsc{Success} has a 95\% interval of [\HumanBaselineCILo, \HumanBaselineCIHi].}
\label{tab:human_baseline}
\end{table}

Human performance establishes substantial headroom on the benchmark while
localizing it to the culturally grounded dimensions.
The leading model is within $4.7$ points of human Visual coverage, whereas the
gaps on Identity and Knowledge are $19.4$ and $20.3$ points, respectively.
This contrast reinforces the central finding that progress in description has
outpaced progress in culturally grounded interpretation.

\section{Evaluation and Diagnostic Analyses}
\label{app:evaldetails}

\subsection{Inference and Judging Protocol}
\label{app:prompt}

The standard prompt used for all models is reproduced below.

\paragraph{Decoding.}
For stochastic decoding, we set temperature to 1.0 and top-$p$ to 0.95;
locally served open-weight models use greedy decoding.
Each model produces one response per item.
We release the complete inference configurations together with the frozen
responses and judge outputs, allowing every reported result to be recomputed
from fixed artifacts.

\begin{prompttable}{Standard inference prompt}
You are an expert at analyzing multimodal content and memes.

Analyze this image step-by-step:
\begin{enumerate}[leftmargin=4.3mm,itemsep=1pt,topsep=2pt,parsep=0pt]
  \item \textbf{Visual Observation}: Describe the visual elements and text
  (OCR) you see in detail.
  \item \textbf{Identity Recognition}: Identify the specific entities,
  characters, origin works, or public figures involved. Use proper nouns
  (character names, work titles), not generic descriptions.
  \item \textbf{Knowledge Retrieval}: Explain the cultural or factual
  background knowledge needed to understand this meme. What would an outsider
  need to know?
  \item \textbf{Cultural Decoding}: Explain the underlying joke, irony, or
  metaphor. How do the visual elements and cultural context combine to create
  humor?
\end{enumerate}
\end{prompttable}

For Chinese memes, an equivalent Chinese prompt is used.

The CoT condition replaces the standard prompt with the reasoning-oriented
prompt shown next.

\begin{prompttable}{CoT inference prompt}
You are an expert at analyzing multimodal content and memes.

Look at this meme image carefully. Think through your
analysis step-by-step, showing your reasoning process as you go.

Start by describing what you see, then work through identifying who or what is
depicted, recall relevant cultural background, and finally explain how the
humor works.

Think aloud as you analyze: show your reasoning, not just your conclusions. If
you're uncertain about something, say so and explain your reasoning for your
best guess.

After your step-by-step analysis, provide a brief summary of:
\begin{itemize}[leftmargin=4.3mm,itemsep=1pt,topsep=2pt,parsep=0pt]
  \item What entities are involved (use proper nouns)
  \item What cultural knowledge is needed
  \item Why it's funny
\end{itemize}
\end{prompttable}

\paragraph{Dual-judge checklist.}
\label{app:judge_prompt}

Two LLM judges (Gemini-3.1-Pro and GPT-5.1) independently evaluate each candidate response against the ground-truth annotation.
Each judge receives the full GT JSON (\texttt{visual}, \texttt{identity}, \texttt{knowledge}, \texttt{reasoning}, and \texttt{eval\_checklist}) along with the candidate response.
The core instructions are reproduced below.

\begin{prompttable}{Dual-judge checklist prompt}
Evaluate the Candidate Response across 4 dimensions defined in the GT's
\texttt{eval\_checklist}. Each dimension contains an array of checklist items
--- you must verify every item in the array.

\textbf{Aggregation Rule.} A dimension scores 1 ONLY if ALL checklist items in
that dimension are satisfied. If any single item fails, the entire dimension
scores 0.

\textbf{Visual Accuracy.} Verify each item in
\texttt{eval\_checklist.visual}. Does the candidate correctly describe key
visual elements and OCR text?

\textbf{Identity Recognition (STRICT).} Verify each item in
\texttt{eval\_checklist.identity}. The candidate MUST use the exact entity
names (e.g., ``Conan Edogawa'' not ``a boy with glasses'', ``Thanos'' not ``a
purple titan''). Generic descriptions always fail.

\textbf{Knowledge Grounding.} Verify each item in
\texttt{eval\_checklist.knowledge}. Does the candidate demonstrate relevant
encyclopedic knowledge? Semantic equivalence is sufficient; exact
wording is not required.

\textbf{Reasoning Logic.} Verify each item in
\texttt{eval\_checklist.reasoning}. Does the candidate explain how trigger
$\to$ reference $\to$ humor mechanism works? Semantic equivalence is
sufficient.
\end{prompttable}

Each judge outputs a binary vector $[V, I, K, R]$ and a checklist score $V{+}I{+}K{+}R$.
A response counts as \textsc{Success} only when all four dimensions pass under both judges (dual-judge intersection).
Both judges are queried with their provider's default decoding parameters and score each response exactly once.

\subsection{Robustness, Sensitivity, and Diagnostic Breakdowns}
\label{app:judge_robustness}

Both judges scored all 26 vanilla runs independently over the full benchmark
($n{=}1{,}253$ each) using the same checklist and rubric version.
Table~\ref{tab:judge_robustness} reports the resulting single-judge and
intersection leaderboards.
\begin{table*}[t!]
\centering
\scriptsize
\setlength{\tabcolsep}{1.2pt}
\setlength{\aboverulesep}{0.25ex}
\setlength{\belowrulesep}{0.25ex}
\renewcommand{\arraystretch}{0.90}
\captionsetup{skip=3pt}
\begin{tabular*}{\textwidth}{@{\extracolsep{\fill}}lrrrrrrrrrrrrrrr@{}}
\toprule
\textbf{Model} & \multicolumn{5}{c}{\textbf{Gemini-3.1-Pro only}} & \multicolumn{5}{c}{\textbf{GPT-5.1 only}} & \multicolumn{5}{c}{\textbf{Intersection}} \\
\cmidrule(lr){2-6}\cmidrule(lr){7-11}\cmidrule(l){12-16}
& \textsc{Succ.} & \textbf{V} & \textbf{I} & \textbf{K} & \textbf{R} & \textsc{Succ.} & \textbf{V} & \textbf{I} & \textbf{K} & \textbf{R} & \textsc{Succ.} & \textbf{V} & \textbf{I} & \textbf{K} & \textbf{R} \\
\midrule
\multicolumn{16}{c}{\emph{Commercial Models}} \\[-0.5pt]
Gemini-3.1-Pro & \textbf{62.8} & \textbf{93.5} & \textbf{78.5} & \textbf{70.3} & \textbf{79.1} & \textbf{72.7} & \textbf{95.6} & \textbf{79.1} & \textbf{85.7} & \textbf{85.7} & \textbf{60.3} & \textbf{92.3} & \textbf{74.6} & \textbf{69.7} & \textbf{76.7} \\
Gemini-3-Flash & 57.9 & 89.6 & 74.5 & 66.4 & 74.5 & 63.0 & 90.5 & 71.8 & 79.2 & 78.5 & 53.7 & 86.0 & 68.7 & 65.5 & 70.3 \\
Qwen3.6-Plus & 31.0 & 70.2 & 52.3 & 44.0 & 50.9 & 40.4 & 78.1 & 52.8 & 61.9 & 63.0 & 28.4 & 65.1 & 47.2 & 42.8 & 47.7 \\
Qwen3.6-Flash & 15.5 & 63.0 & 35.9 & 26.6 & 32.9 & 20.7 & 66.3 & 33.8 & 45.7 & 45.1 & 12.3 & 53.2 & 28.6 & 25.7 & 29.0 \\
GPT-5.6-Sol & 42.5 & 75.7 & 61.9 & 55.6 & 63.0 & 55.1 & 83.9 & 63.8 & 75.7 & 74.8 & 40.9 & 72.5 & 58.7 & 55.1 & 60.7 \\
GPT-5.6-Terra & 29.8 & 65.7 & 49.4 & 42.4 & 48.2 & 40.1 & 76.9 & 51.6 & 65.7 & 67.4 & 27.8 & 61.9 & 45.4 & 42.0 & 46.9 \\
GPT-5.6-Luna & 19.3 & 56.1 & 38.5 & 31.6 & 36.3 & 30.0 & 69.8 & 42.6 & 55.9 & 57.5 & 18.0 & 52.1 & 34.9 & 31.3 & 35.0 \\
Claude-Opus-5 & 40.8 & 77.2 & 54.7 & 51.2 & 60.0 & 50.0 & 87.8 & 56.5 & 75.3 & 73.4 & 38.3 & 74.7 & 50.4 & 50.9 & 57.9 \\
Claude-Sonnet-5 & 19.0 & 65.4 & 37.7 & 30.7 & 38.5 & 28.1 & 73.1 & 40.1 & 52.1 & 52.1 & 17.8 & 59.2 & 33.3 & 30.2 & 35.6 \\
\midrule
\multicolumn{16}{c}{\emph{Open-Source Models}} \\[-0.5pt]
Kimi-K2.5 & 41.2 & 74.8 & 64.5 & 55.5 & 62.3 & 52.6 & 83.3 & 65.8 & 75.7 & 73.7 & 37.9 & 71.1 & 59.2 & 54.5 & 58.7 \\
MiMo-V2.5 & 15.7 & 56.1 & 34.1 & 27.9 & 33.1 & 22.3 & 62.0 & 35.8 & 48.8 & 44.5 & 13.7 & 47.2 & 30.5 & 26.9 & 28.3 \\
MiMo-V2-Omni & 16.4 & 57.9 & 34.2 & 27.5 & 32.7 & 21.9 & 61.1 & 35.1 & 49.2 & 45.6 & 14.3 & 48.1 & 29.7 & 26.4 & 28.8 \\
Qwen3-VL-235B-A22B-Instruct & 16.8 & 63.0 & 37.1 & 26.7 & 33.1 & 22.7 & 64.3 & 35.7 & 48.3 & 45.1 & 14.5 & 53.2 & 30.6 & 26.1 & 27.3 \\
Qwen3-VL-235B-A22B-Thinking & 13.4 & 57.7 & 32.0 & 23.8 & 28.6 & 20.1 & 65.9 & 32.6 & 44.8 & 41.4 & 12.0 & 50.2 & 27.5 & 22.9 & 24.7 \\
Qwen3-VL-32B-Instruct & 10.3 & 59.7 & 26.8 & 19.8 & 24.7 & 14.0 & 60.0 & 23.6 & 38.5 & 34.7 & 8.4 & 47.6 & 20.1 & 19.1 & 20.0 \\
Qwen3-VL-32B-Thinking & 8.9 & 53.3 & 25.1 & 18.9 & 23.3 & 14.2 & 59.7 & 24.5 & 36.7 & 36.0 & 7.7 & 44.5 & 20.5 & 17.9 & 19.7 \\
Qwen3-VL-30B-A3B-Instruct & 5.8 & 52.9 & 23.2 & 14.0 & 15.5 & 7.3 & 51.2 & 17.4 & 26.8 & 23.5 & 3.3 & 39.7 & 14.8 & 12.2 & 11.3 \\
Qwen3-VL-30B-A3B-Thinking & 7.0 & 52.3 & 24.3 & 15.8 & 19.2 & 11.7 & 57.1 & 22.8 & 33.0 & 31.0 & 5.8 & 41.6 & 18.3 & 14.7 & 15.8 \\
Qwen3-VL-8B-Instruct & 4.2 & 50.0 & 19.9 & 10.9 & 12.6 & 6.5 & 44.5 & 15.6 & 23.8 & 20.8 & 3.3 & 34.0 & 13.2 & 9.3 & 9.3 \\
Qwen3-VL-8B-Thinking & 4.1 & 43.2 & 18.6 & 10.7 & 13.1 & 7.7 & 47.2 & 17.2 & 24.2 & 24.2 & 3.4 & 32.1 & 14.0 & 10.0 & 9.9 \\
InternVL3.5-38B & 2.8 & 46.8 & 17.3 & 8.6 & 12.1 & 6.5 & 52.9 & 15.1 & 22.3 & 22.3 & 2.6 & 35.6 & 12.5 & 7.8 & 8.2 \\
InternVL3.5-30B-A3B & 1.7 & 37.3 & 14.2 & 6.7 & 9.0 & 3.0 & 44.4 & 10.9 & 17.2 & 17.1 & 1.1 & 26.4 & 8.8 & 5.6 & 6.2 \\
InternVL3.5-20B-A4B & 1.6 & 33.8 & 13.2 & 6.3 & 7.9 & 2.6 & 37.2 & 10.1 & 15.4 & 13.5 & 1.1 & 22.0 & 8.5 & 5.4 & 5.3 \\
InternVL3.5-14B & 1.5 & 34.8 & 14.7 & 7.3 & 9.3 & 4.1 & 43.9 & 11.7 & 17.6 & 15.8 & 1.1 & 26.0 & 9.3 & 6.0 & 6.1 \\
InternVL3.5-8B & 1.4 & 29.1 & 10.8 & 5.7 & 6.5 & 3.4 & 37.3 & 11.3 & 15.9 & 15.3 & 1.0 & 21.1 & 8.3 & 5.2 & 4.9 \\
LLaVA-OV-8B-RL & 0.6 & 24.7 & 13.5 & 4.5 & 4.1 & 3.5 & 35.2 & 12.5 & 15.9 & 14.3 & 0.4 & 18.8 & 9.0 & 4.2 & 3.2 \\
\bottomrule
\end{tabular*}
\caption{Vanilla leaderboard under each judge alone and under their intersection (\%, overall split, $n{=}1{,}253$). Bold marks the best value within each judge block. The two single-judge rankings correlate at Spearman $\rho{=}0.99$, and each correlates with the intersection at $\rho{\geq}0.99$; Gemini-3.1-Pro ranks first under all three evaluation views.}
\label{tab:judge_robustness}
\end{table*}

The model ordering is highly stable across evaluation views: the two
single-judge leaderboards correlate at Spearman $\rho{=}0.99$, and each
correlates with the intersection at $\rho{\geq}0.99$.
Gemini-3.1-Pro ranks first under all three views.
Together with the human calibration in Table~\ref{tab:judge_human}, this
agreement supports using the dual-judge intersection as a stable comparative
measure.


\paragraph{VIKR diagnostic breakdowns.}
\label{app:vikrdetails}

\textbf{Response-state distribution.}
\label{app:states}
Table~\ref{tab:vikr_states} reports the full distribution over the 16 VIKR states for each model with a complete judge pass.

\begin{table*}[t]
\centering
\scriptsize
\setlength{\tabcolsep}{2.6pt}
\begin{tabular*}{\textwidth}{@{\extracolsep{\fill}}lrrrrrrrrrrrrrrrr@{}}
\toprule
\textbf{Model} & \texttt{0000} & \texttt{0001} & \texttt{0010} & \texttt{0011} & \texttt{0100} & \texttt{0101} & \texttt{0110} & \texttt{0111} & \texttt{1000} & \texttt{1001} & \texttt{1010} & \texttt{1011} & \texttt{1100} & \texttt{1101} & \texttt{1110} & \texttt{1111} \\
\midrule
Gemini-3.1-Pro & 3.4 & 0.4 & 0.2 & 0.5 & 1.0 & 0.2 & 0.7 & 1.4 & 12.1 & 3.6 & 1.3 & 3.9 & 3.2 & 6.5 & 1.4 & 60.3 \\
Gemini-3-Flash & 7.1 & 0.0 & 0.2 & 0.7 & 1.7 & 0.8 & 1.4 & 2.0 & 13.4 & 3.2 & 1.4 & 5.1 & 3.4 & 4.8 & 1.0 & 53.8 \\
GPT-5.6-Sol & 15.7 & 1.4 & 1.4 & 1.0 & 2.1 & 0.3 & 1.6 & 3.8 & 12.9 & 3.7 & 1.0 & 4.2 & 3.3 & 5.3 & 1.2 & 40.9 \\
Claude-Opus-5 & 16.5 & 1.5 & 1.2 & 1.0 & 1.0 & 0.4 & 0.8 & 2.7 & 17.8 & 5.5 & 0.8 & 5.2 & 3.0 & 3.3 & 0.9 & 38.4 \\
Kimi-K2.5 & 14.2 & 1.1 & 1.8 & 1.0 & 3.8 & 1.3 & 1.3 & 4.2 & 12.6 & 3.6 & 1.1 & 5.2 & 4.3 & 4.4 & 2.0 & 38.0 \\
Qwen3.6-Plus & 22.3 & 1.1 & 1.0 & 1.4 & 2.7 & 1.3 & 1.2 & 3.3 & 16.1 & 4.5 & 1.7 & 4.3 & 5.2 & 3.6 & 1.6 & 28.7 \\
GPT-5.6-Terra & 24.6 & 1.8 & 1.2 & 1.6 & 3.5 & 0.7 & 1.6 & 3.1 & 15.9 & 4.2 & 1.1 & 4.3 & 3.9 & 3.5 & 1.3 & 27.8 \\
GPT-5.6-Luna & 33.8 & 2.2 & 1.8 & 1.2 & 3.4 & 0.8 & 1.6 & 3.0 & 18.0 & 3.9 & 1.1 & 3.1 & 4.0 & 2.6 & 1.4 & 18.0 \\
Claude-Sonnet-5 & 30.3 & 1.3 & 1.1 & 1.0 & 2.6 & 0.9 & 1.4 & 2.1 & 21.9 & 5.3 & 1.3 & 4.5 & 4.8 & 2.7 & 0.9 & 17.8 \\
Qwen3-VL-235B-A22B-Instruct & 35.4 & 1.5 & 1.1 & 0.5 & 4.4 & 0.6 & 1.7 & 1.4 & 21.8 & 3.0 & 2.6 & 3.3 & 4.6 & 2.5 & 1.0 & 14.6 \\
MiMo-V2-Omni & 40.5 & 1.4 & 1.2 & 1.0 & 3.5 & 1.3 & 1.2 & 1.8 & 17.6 & 3.5 & 1.6 & 3.5 & 3.8 & 2.1 & 1.8 & 14.3 \\
MiMo-V2.5 & 39.4 & 1.9 & 1.4 & 0.9 & 4.0 & 0.8 & 1.8 & 2.5 & 18.0 & 2.3 & 1.6 & 3.9 & 4.3 & 2.3 & 1.0 & 13.8 \\
Qwen3.6-Flash & 35.5 & 2.2 & 1.2 & 1.2 & 3.0 & 0.4 & 0.8 & 2.3 & 21.1 & 3.6 & 2.0 & 4.5 & 5.8 & 2.6 & 1.4 & 12.3 \\
Qwen3-VL-235B-A22B-Thinking & 38.2 & 1.8 & 1.7 & 0.8 & 4.0 & 0.5 & 0.9 & 1.7 & 22.0 & 3.3 & 1.8 & 2.7 & 5.1 & 2.1 & 1.4 & 12.0 \\
Qwen3-VL-32B-Instruct & 43.4 & 1.4 & 1.5 & 1.0 & 2.8 & 0.6 & 0.8 & 0.7 & 24.4 & 2.7 & 1.9 & 3.5 & 3.7 & 1.8 & 1.4 & 8.4 \\
Qwen3-VL-32B-Thinking & 44.2 & 1.8 & 1.3 & 0.8 & 3.8 & 0.6 & 0.7 & 1.9 & 23.9 & 2.6 & 1.9 & 2.8 & 3.3 & 1.8 & 1.0 & 7.7 \\
Qwen3-VL-30B-A3B-Thinking & 48.5 & 1.0 & 1.8 & 0.7 & 4.2 & 0.2 & 0.8 & 1.1 & 23.5 & 2.6 & 1.4 & 2.2 & 3.1 & 2.2 & 0.9 & 5.8 \\
Qwen3-VL-8B-Thinking & 57.8 & 1.3 & 1.9 & 0.8 & 3.8 & 0.2 & 0.7 & 0.5 & 20.0 & 1.6 & 1.1 & 1.1 & 3.9 & 1.1 & 0.7 & 3.4 \\
Qwen3-VL-8B-Instruct & 58.7 & 1.0 & 0.9 & 0.4 & 2.9 & 0.3 & 0.9 & 0.8 & 22.2 & 1.3 & 1.0 & 1.3 & 3.3 & 1.0 & 0.8 & 3.3 \\
Qwen3-VL-30B-A3B-Instruct & 52.4 & 0.7 & 1.4 & 0.8 & 2.7 & 0.6 & 0.5 & 1.0 & 24.4 & 1.7 & 1.8 & 1.8 & 3.8 & 1.4 & 1.6 & 3.3 \\
InternVL3.5-38B & 57.4 & 1.2 & 0.8 & 0.3 & 3.1 & 0.5 & 0.6 & 0.5 & 24.2 & 1.7 & 1.1 & 0.8 & 3.5 & 0.6 & 1.0 & 2.6 \\
InternVL3.5-30B-A3B & 66.6 & 1.4 & 1.1 & 0.5 & 2.7 & 0.7 & 0.2 & 0.3 & 18.9 & 1.0 & 1.0 & 0.8 & 2.7 & 0.4 & 0.6 & 1.1 \\
InternVL3.5-20B-A4B & 70.9 & 0.7 & 1.0 & 0.3 & 3.7 & 0.2 & 0.6 & 0.5 & 15.9 & 1.1 & 0.7 & 0.8 & 1.6 & 0.5 & 0.3 & 1.1 \\
InternVL3.5-14B & 66.6 & 1.2 & 0.8 & 0.6 & 3.0 & 0.4 & 0.9 & 0.5 & 18.8 & 0.9 & 0.9 & 0.9 & 2.4 & 0.6 & 0.4 & 1.1 \\
InternVL3.5-8B & 72.1 & 0.8 & 1.5 & 0.3 & 3.2 & 0.2 & 0.6 & 0.2 & 14.8 & 1.1 & 0.4 & 0.6 & 2.0 & 0.6 & 0.5 & 1.0 \\
LLaVA-OV-8B-RL & 74.8 & 0.7 & 1.4 & 0.1 & 3.5 & 0.0 & 0.6 & 0.2 & 12.4 & 0.6 & 0.6 & 0.4 & 3.0 & 0.8 & 0.5 & 0.4 \\
\bottomrule
\end{tabular*}
\caption{Distribution over the 16 VIKR pass/fail states (\% of scored responses, digits ordered V\,I\,K\,R). Across the 26 Vanilla runs, 16 of 16 states occur in the pooled responses, including states such as \texttt{0001} and \texttt{0010} that a deterministic V--I--K--R chain would not produce. Percentages exclude items that either judge failed to score.}
\label{tab:vikr_states}
\end{table*}

\textbf{Domain and language strata.}
\label{app:strata}
Tables~\ref{tab:domain_breakdown} and~\ref{tab:language_breakdown} report vanilla \textsc{Success} by domain and by language, the latter both overall and restricted to the ACG domain.

\begin{table*}[t!]
\centering
\scriptsize
\setlength{\tabcolsep}{3.2pt}
\setlength{\aboverulesep}{0.25ex}
\setlength{\belowrulesep}{0.25ex}
\renewcommand{\arraystretch}{0.96}
\captionsetup{skip=3pt}
\begin{tabular*}{\textwidth}{@{\extracolsep{\fill}}lrrrrr@{}}
\toprule
\textbf{Model} & \textbf{ACG} & \textbf{Cross-Dom.} & \textbf{Daily} & \textbf{Movies} & \textbf{Public} \\
 & (628) & (266) & (166) & (125) & (68) \\
\midrule
Gemini-3.1-Pro & 52.1 & 65.8 & 69.3 & 69.6 & 75.0 \\
Gemini-3-Flash & 44.0 & 60.5 & 62.6 & 66.4 & 72.1 \\
GPT-5.6-Sol & 32.2 & 42.9 & 51.2 & 54.4 & 64.7 \\
Kimi-K2.5 & 28.0 & 41.0 & 42.8 & 59.2 & 66.2 \\
Claude-Opus-5 & 23.4 & 44.0 & 56.6 & 59.2 & 70.6 \\
Qwen3.6-Plus & 18.6 & 31.2 & 37.4 & 42.4 & 60.3 \\
GPT-5.6-Terra & 17.5 & 32.3 & 38.0 & 38.4 & 60.3 \\
GPT-5.6-Luna & 10.0 & 18.4 & 24.1 & 34.4 & 45.6 \\
Claude-Sonnet-5 & 7.2 & 19.9 & 30.7 & 32.8 & 48.5 \\
MiMo-V2-Omni & 6.7 & 16.9 & 19.9 & 24.0 & 42.6 \\
Qwen3-VL-235B-A22B-Instruct & 6.4 & 15.0 & 23.5 & 25.6 & 45.6 \\
Qwen3.6-Flash & 5.6 & 13.9 & 18.1 & 21.6 & 36.8 \\
MiMo-V2.5 & 4.9 & 18.8 & 21.1 & 24.0 & 38.2 \\
Qwen3-VL-235B-A22B-Thinking & 4.1 & 11.7 & 24.1 & 24.0 & 33.8 \\
Qwen3-VL-32B-Thinking & 3.0 & 7.5 & 12.1 & 15.2 & 26.5 \\
Qwen3-VL-32B-Instruct & 1.9 & 8.3 & 18.1 & 17.6 & 27.9 \\
Qwen3-VL-30B-A3B-Thinking & 1.1 & 6.8 & 10.2 & 12.0 & 23.5 \\
Qwen3-VL-8B-Thinking & 0.6 & 2.3 & 8.4 & 8.0 & 11.8 \\
InternVL3.5-38B & 0.5 & 2.3 & 7.2 & 4.8 & 8.8 \\
InternVL3.5-30B-A3B & 0.5 & 0.4 & 1.8 & 4.0 & 2.9 \\
Qwen3-VL-30B-A3B-Instruct & 0.3 & 5.3 & 4.8 & 6.4 & 13.2 \\
Qwen3-VL-8B-Instruct & 0.3 & 2.6 & 8.4 & 8.0 & 11.8 \\
InternVL3.5-20B-A4B & 0.3 & 0.4 & 1.8 & 3.2 & 5.9 \\
InternVL3.5-14B & 0.3 & 1.1 & 3.0 & 1.6 & 2.9 \\
InternVL3.5-8B & 0.2 & 1.1 & 3.0 & 0.8 & 4.4 \\
LLaVA-OV-8B-RL & 0.0 & 0.4 & 0.6 & 1.6 & 1.5 \\
\bottomrule
\end{tabular*}
\caption{Dual-judge Vanilla \textsc{Success} (\%) by domain, with item counts in parentheses. Unscored responses count as failures. Per-domain V/I/K/R and 95\% bootstrap intervals are released with the analysis code.}
\label{tab:domain_breakdown}

\vspace{0.8em}

\begin{tabular*}{\textwidth}{@{\extracolsep{\fill}}lrrrrrr@{}}
\toprule
& \multicolumn{3}{c}{\textbf{All domains}} & \multicolumn{3}{c}{\textbf{Within ACG}} \\
\cmidrule(lr){2-4}\cmidrule(l){5-7}
\textbf{Model} & \textbf{ZH} & \textbf{EN} & $\boldsymbol{\Delta}$ & \textbf{ZH} & \textbf{EN} & $\boldsymbol{\Delta}$ \\
\midrule
Gemini-3.1-Pro & 49.9 & 76.7 & +26.8 & 47.1 & 64.9 & +17.8 \\
Gemini-3-Flash & 41.8 & 72.6 & +30.8 & 37.9 & 59.8 & +21.9 \\
Claude-Opus-5 & 28.3 & 54.2 & +26.0 & 18.5 & 36.2 & +17.7 \\
GPT-5.6-Sol & 33.1 & 53.4 & +20.3 & 30.4 & 36.8 & +6.4 \\
Kimi-K2.5 & 32.4 & 46.6 & +14.2 & 26.4 & 32.2 & +5.8 \\
GPT-5.6-Terra & 22.3 & 36.5 & +14.2 & 16.7 & 19.5 & +2.8 \\
Qwen3.6-Plus & 25.5 & 33.0 & +7.5 & 17.8 & 20.7 & +2.9 \\
Claude-Sonnet-5 & 12.2 & 26.6 & +14.4 & 5.7 & 10.9 & +5.2 \\
GPT-5.6-Luna & 15.5 & 22.1 & +6.6 & 10.6 & 8.6 & -2.0 \\
Qwen3-VL-235B-A22B-Instruct & 10.2 & 21.4 & +11.3 & 4.4 & 11.5 & +7.1 \\
MiMo-V2.5 & 9.9 & 19.8 & +9.9 & 3.7 & 8.1 & +4.3 \\
MiMo-V2-Omni & 10.8 & 19.8 & +9.0 & 6.0 & 8.6 & +2.7 \\
Qwen3-VL-235B-A22B-Thinking & 9.6 & 15.7 & +6.0 & 3.1 & 6.9 & +3.8 \\
Qwen3.6-Flash & 10.6 & 15.1 & +4.5 & 4.8 & 7.5 & +2.6 \\
Qwen3-VL-32B-Instruct & 5.9 & 12.4 & +6.5 & 1.1 & 4.0 & +2.9 \\
Qwen3-VL-32B-Thinking & 5.5 & 11.1 & +5.7 & 1.8 & 6.3 & +4.6 \\
Qwen3-VL-30B-A3B-Thinking & 3.6 & 9.3 & +5.6 & 0.2 & 3.5 & +3.2 \\
Qwen3-VL-8B-Instruct & 1.9 & 5.4 & +3.4 & 0.2 & 0.6 & +0.3 \\
InternVL3.5-38B & 1.4 & 4.5 & +3.1 & 0.0 & 1.7 & +1.7 \\
Qwen3-VL-8B-Thinking & 2.7 & 4.3 & +1.6 & 0.2 & 1.7 & +1.5 \\
Qwen3-VL-30B-A3B-Instruct & 2.7 & 4.1 & +1.4 & 0.0 & 1.1 & +1.1 \\
InternVL3.5-30B-A3B & 0.7 & 1.9 & +1.2 & 0.2 & 1.1 & +0.9 \\
InternVL3.5-8B & 0.7 & 1.6 & +1.0 & 0.0 & 0.6 & +0.6 \\
InternVL3.5-20B-A4B & 1.2 & 1.0 & -0.1 & 0.2 & 0.6 & +0.3 \\
InternVL3.5-14B & 1.2 & 1.0 & -0.1 & 0.2 & 0.6 & +0.3 \\
LLaVA-OV-8B-RL & 0.1 & 0.8 & +0.7 & 0.0 & 0.0 & +0.0 \\
\bottomrule
\end{tabular*}
\caption{Dual-judge Vanilla \textsc{Success} (\%) by language, over all domains and restricted to ACG. Unscored responses count as failures. The within-ACG comparison controls domain composition more closely; both views characterize the two sampled language-associated online-cultural ecosystems.}
\label{tab:language_breakdown}
\end{table*}

\paragraph{Metric sensitivity.}
\label{app:additional_analyses}
\label{app:metric_sensitivity}

The primary \textsc{Success} metric asks whether an explanation covers all four VIKR dimensions.
We define a complementary item-level sensitivity metric
\begin{equation}
  \mathrm{Success}^{3/4}_i
  = \mathbf{1}\!\left\{V_i+I_i+K_i+R_i \geq 3\right\},
\end{equation}
which tests whether conclusions depend on requiring complete four-dimensional coverage.

Table~\ref{tab:metric_sensitivity} reports this sensitivity analysis for the
26-model Vanilla leaderboard under the same dual-judge and full-denominator
protocol as the primary results.
The relaxed criterion raises absolute rates by 1.9--15.8 percentage points
while preserving the model ordering closely (Spearman $\rho{=}0.99$), so the
leaderboard's capability gradient does not depend on the strict conjunction.
\par\medskip
\noindent\begin{minipage}{\columnwidth}
\centering
\fontsize{5.8}{6.2}\selectfont
\setlength{\tabcolsep}{3.0pt}
\setlength{\aboverulesep}{0.25ex}
\setlength{\belowrulesep}{0.25ex}
\renewcommand{\arraystretch}{0.82}
\captionsetup{font=footnotesize,skip=2pt}
\begin{tabular}{@{}lrrr@{}}
\toprule
\textbf{Model} & \textbf{4/4} & \textbf{3/4} & $\boldsymbol{\Delta}$ \\
\midrule
Gemini-3.1-Pro & 60.3 & 73.5 & +13.2 \\
Gemini-3-Flash & 53.7 & 66.6 & +12.9 \\
GPT-5.6-Sol & 40.9 & 55.5 & +14.5 \\
Claude-Opus-5 & 38.3 & 50.4 & +12.0 \\
Kimi-K2.5 & 37.9 & 53.7 & +15.8 \\
Qwen3.6-Plus & 28.4 & 41.2 & +12.8 \\
GPT-5.6-Terra & 27.8 & 40.0 & +12.2 \\
GPT-5.6-Luna & 18.0 & 28.2 & +10.1 \\
Claude-Sonnet-5 & 17.8 & 28.1 & +10.3 \\
Qwen3-VL-235B-A22B-Instruct & 14.5 & 22.8 & +8.2 \\
MiMo-V2-Omni & 14.3 & 23.5 & +9.2 \\
MiMo-V2.5 & 13.7 & 23.5 & +9.7 \\
Qwen3.6-Flash & 12.3 & 23.1 & +10.8 \\
Qwen3-VL-235B-A22B-Thinking & 12.0 & 19.8 & +7.8 \\
Qwen3-VL-32B-Instruct & 8.4 & 15.7 & +7.3 \\
Qwen3-VL-32B-Thinking & 7.7 & 15.0 & +7.3 \\
Qwen3-VL-30B-A3B-Thinking & 5.8 & 12.1 & +6.3 \\
Qwen3-VL-8B-Thinking & 3.4 & 6.7 & +3.4 \\
Qwen3-VL-30B-A3B-Instruct & 3.3 & 9.1 & +5.8 \\
Qwen3-VL-8B-Instruct & 3.3 & 7.1 & +3.8 \\
InternVL3.5-38B & 2.6 & 5.6 & +3.0 \\
InternVL3.5-30B-A3B & 1.1 & 3.2 & +2.1 \\
InternVL3.5-20B-A4B & 1.1 & 3.2 & +2.1 \\
InternVL3.5-14B & 1.1 & 3.5 & +2.4 \\
InternVL3.5-8B & 1.0 & 3.0 & +1.9 \\
LLaVA-OV-8B-RL & 0.4 & 2.3 & +1.9 \\
\bottomrule
\end{tabular}
\captionof{table}{Metric sensitivity for the 26-model Vanilla leaderboard under the dual-judge intersection (\%). \textbf{4/4} is the primary \textsc{Success} criterion; \textbf{3/4} accepts any response covering at least three VIKR dimensions. $\Delta$ is the absolute increase under the relaxed aggregation. The denominator remains all 1,253 items, so unscored responses count as failures.}
\label{tab:metric_sensitivity}
\end{minipage}
\par\medskip

\section{KAR and CultureBase}
\label{app:kardetails}

This appendix gives the formal definition of the entity-guided retrieval condition shown in Figure~\ref{fig:kar} and the construction of CultureBase.

\subsection{Retrieval Pipeline}
\label{app:kar_formal}

\paragraph{Stage 1, LVLM extraction.}
Given a meme image $x_i$, an extractor prompt asks the LVLM to produce
\begin{equation}
  z_i = (o_i, h_i, \mathcal{H}_i, \mathcal{Q}^{\mathrm{vlm}}_i),
\end{equation}
where $o_i$ is OCR text, $h_i$ is a short visual cue phrase, $\mathcal{H}_i=\{(n_\ell,s_\ell,c_\ell)\}_{\ell=1}^{L_i}$ is a set of entity hypotheses (name, source, confidence), and $\mathcal{Q}^{\mathrm{vlm}}_i$ is a set of model-suggested web queries.

\paragraph{Stage 2, CultureBase retrieval.}
To ground the extracted entity hypotheses, KAR builds a three-route query set from OCR text, visual cue phrase, and entity names:
\begin{equation}
\begin{aligned}
\mathcal{R}_i ={}&
  \{o_i\}
  \cup \{h_i\} \\
  &\cup \{n_\ell : (n_\ell,s_\ell,c_\ell)\in\mathcal{H}_i,\;
  c_\ell\in\{\mathrm{medium},\mathrm{high}\}\}.
\end{aligned}
\end{equation}
Each CultureBase entity is scored by its best match across routes:
\begin{equation}
  \mathrm{score}_{ij}=\max_{r\in\mathcal{R}_i} g(r)^\top u_j,
\end{equation}
where $g(\cdot)$ is the BGE-M3 embedding function and $u_j$ is the entity embedding.
The top-$K$ entities above a similarity threshold are retained:
\begin{equation}
\begin{aligned}
  \widehat{\mathcal{C}}_i
  ={}& \{e_j : j \in {} \\
  &\quad \operatorname{TopK}_{j:\,\mathrm{score}_{ij}\geq\tau}
  (\mathrm{score}_{ij})\}, \\
  &\qquad K{=}5,\; \tau{=}0.5.
\end{aligned}
\end{equation}

\paragraph{Stage 3, entity-guided web search.}
Retrieved entities are converted into search queries via a template $\psi(e_j)$ (e.g., ``\texttt{<name> <source> meme origin}'').
The final query set combines model-suggested and entity-grounded queries under a shared budget $B{=}5$:
\begin{equation}
  \mathcal{Q}^{\mathrm{kar}}_i
    = \operatorname{Budget}_{B}
    \left(\mathcal{Q}^{\mathrm{vlm}}_i
    \cup \{\psi(e_j): e_j\in\widehat{\mathcal{C}}_i\}\right).
\end{equation}
Each query is executed with a web-search backend (Tavily), yielding a combined evidence set $\mathcal{S}_i$.

\paragraph{Stage 4, grounded reasoning.}
The LVLM receives the original image together with the retrieved evidence $\mathcal{S}_i$ and produces the final explanation.
In the Vanilla and CoT conditions this evidence set is empty, and in Search-CoT it is populated without the CultureBase stage.

\subsection{CultureBase}
\label{app:culturebase}

CultureBase is a curated knowledge base of ${\sim}$5{,}000 cultural entities crawled from Moegirl (Chinese ACG encyclopedia) and KnowYourMeme (English meme database).
Each record stores the entity's name, aliases, source work, character traits, visual description, and cultural usage patterns.
Entities are embedded with bge-m3~\citep{chen2024bge} for multilingual cosine-similarity retrieval.

These records provide entity anchors for the KAR pipeline
(\S\ref{sec:kar_method}); the retrieved names, aliases, and source works are
then used to guide web searches for cultural background.

\section{Responsible Use and Dataset Card}
\label{app:discussion}

\subsection{Scope and Responsible Use}

\paragraph{Research scope.}
MemeBench is ACG-centered by design.
ACG communities are where meme conventions are densest and where the knowledge needed to decode an image is most clearly external to the image itself, which is the regime this benchmark is built to measure.
This concentrated setting makes long-tail entities, source works, and
community conventions central to the task, while the non-ACG half of the
benchmark tests the same evaluation framework across adjacent online-cultural
domains.
We release a frozen, dated snapshot so that all reported results remain
reproducible as online culture evolves.

\paragraph{Data governance.}
Meme images are sourced from publicly accessible platforms (Bilibili, Reddit, ImgFlip).
We release annotations, metadata, evaluation code, and CultureBase under CC~BY-NC-SA~4.0.
Meme images are hosted alongside annotations on the dataset repository for reviewer access and reproducibility.

\paragraph{Ethical considerations.}
Memes frequently contain stereotypes, political commentary, and culturally sensitive content.
MemeBench excludes explicitly hateful content, and some retained memes carry cultural stereotypes that circulate in their source communities.
Required cultural knowledge is defined relative to the source community of each meme, which is what makes the target decodable and auditable.
The benchmark is designed for diagnostic evaluation.

\subsection{Dataset Card}
\label{app:datacard}

\paragraph{Dataset name.} MemeBench v1

\paragraph{Size.} 1,253 memes (768 Chinese, 485 English) after quality filtering from an initial pool of 1,500. All experiments use the full set; null responses are scored as failures, not excluded.

\paragraph{Task.} Open-ended meme explanation.

\paragraph{Annotation schema.} Four-layer VIKR: Visual, Identity, Knowledge, Reasoning. Per-dimension binary evaluation checklists.

\paragraph{Sources.} Chinese: Bilibili. English: Reddit, ImgFlip. All from publicly accessible content.

\paragraph{Annotation process.} Human-written explanations (\texttt{checked\_gt}) $\to$ model-assisted structured annotation $\to$ automated validation and targeted review $\to$ two blind explanations on a stratified subset $\to$ independent adjudication and reference/checklist freeze.

\paragraph{Intended use.} Diagnostic evaluation of LVLM cultural knowledge. Not intended for training.

\paragraph{Scope.} ACG-centered by design, with adjacent online-cultural domains represented in the same evaluation framework. The release is a dated snapshot, and automatic evaluation is calibrated against human labels and aggregated across two judges.

\paragraph{License.} Annotations, metadata, evaluation code, and CultureBase are released under CC~BY-NC-SA~4.0. Meme images are hosted on the dataset repository. A takedown protocol is included in the repository.

\clearpage
\onecolumn
\section{Qualitative VIKR Cases}
\label{app:vikr_case_table}

Table~\ref{tab:vikr_case_keywords} shows five representative cases in a compact table format.
The top two rows are English memes, and the bottom three rows are Chinese memes.
Each annotation cell expands the original VIKR fields from the MemeBench annotations.

\begingroup
\fontsize{5}{5.2}\selectfont
\setlength{\tabcolsep}{5pt}
\renewcommand{\arraystretch}{1.00}
\begin{longtable}{@{}p{0.33\textwidth}p{0.58\textwidth}@{}}
\caption{Raw qualitative VIKR table for two English and three Chinese meme cases. Each row presents the meme image and the original structured Visual, Identity, Knowledge, and Reasoning annotation fields. Annotation text is reproduced verbatim; the translation and romanization glosses under each Chinese image are added here for readability and are not part of the annotation.}
\label{tab:vikr_case_keywords}\\
\toprule
\textbf{Meme} & \textbf{Raw VIKR annotation} \\
\midrule
\endfirsthead
\toprule
\textbf{Meme} & \textbf{Raw VIKR annotation} \\
\midrule
\endhead
\begin{minipage}[t]{0.33\textwidth}\vspace{0pt}\centering\includegraphics[width=\linewidth]{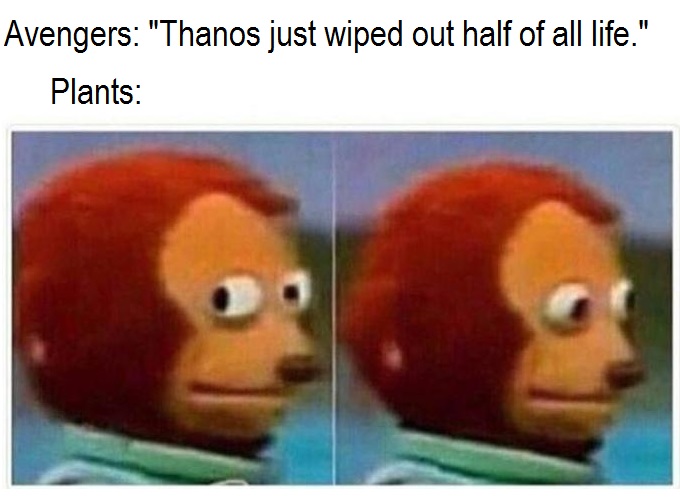}\end{minipage} &
\begin{minipage}[t]{0.58\textwidth}\vspace{0pt}
\raggedright
\{\newline
\hspace*{0.8em}\textcolor{blue}{\textbf{"visual"}}: \{\newline
\hspace*{1.6em}"description": "The image consists of black text at the top and a two-panel image below. The top text is a dialogue. The two-panel image shows a fuzzy puppet character with brown hair and a tan face. In the left panel, the puppet's large eyes are looking sideways towards the viewer. In the right panel, the puppet's eyes have shifted to look forward, avoiding eye contact. The puppet's expression is neutral but appears nervous or awkward due to the eye movement.",\newline
\hspace*{1.6em}"ocr": "Avengers: Thanos just wiped out half of all life. / Plants:",\newline
\hspace*{1.6em}"entities": ["e1: A fuzzy puppet character with brown hair and a tan face, looking sideways in the first panel and then looking away in the second panel, conveying an awkward or nervous expression."]\newline
\hspace*{0.8em}\},\newline
\hspace*{0.8em}\textcolor{violet}{\textbf{"identity"}}: \{\newline
\hspace*{1.6em}"entities": ["e1: name: Monkey Puppet / Pedro the Monkey; source: Okiku Naru Ko."]\newline
\hspace*{0.8em}\},\newline
\hspace*{0.8em}\textcolor{orange}{\textbf{"knowledge"}}: \{\newline
\hspace*{1.6em}"facts": ["In the Marvel Cinematic Universe, Thanos's snap is an event that eliminates half of all living beings in the universe.", "The films showing the aftermath of Thanos's snap focused on the disappearance of humans and animals, largely ignoring the effect on plant life.", "The Monkey Puppet reaction image, often called Awkward Look Monkey, is widely used in online culture to convey feelings of discomfort, guilt, or wanting to avoid attention."],\newline
\hspace*{1.6em}"cultural\_code": "Monkey Puppet meme (Awkward Look Monkey)"\newline
\hspace*{0.8em}\},\newline
\hspace*{0.8em}\textcolor{green}{\textbf{"reasoning"}}: \{\newline
\hspace*{1.6em}"trigger": "The juxtaposition of the text stating half of all life being wiped out, followed by Plants: and the awkward monkey puppet looking away.",\newline
\hspace*{1.6em}"reference": "The realization that plants are technically life and should have been affected by Thanos's snap, but the movies completely ignored this, so plants are humorously depicted as awkwardly staying quiet to avoid drawing attention to their survival.",\newline
\hspace*{1.6em}"core\_message": "The meme points out the logical flaw in Avengers: Infinity War where Thanos wiped out half of all life but plants seemingly survived unharmed, humorously personifying plants as awkwardly trying not to draw attention to this oversight.",\newline
\hspace*{1.6em}"logic": "RV",\newline
\hspace*{1.6em}"logic\_detail": "Rule Violation (logical plot hole in the movie's premise) combined with Identity Collision (personifying plants as having awkward human reactions)."\newline
\hspace*{0.8em}\}\newline
\}
\end{minipage} \\
\midrule
\begin{minipage}[t]{0.33\textwidth}\vspace{0pt}\centering\includegraphics[width=\linewidth]{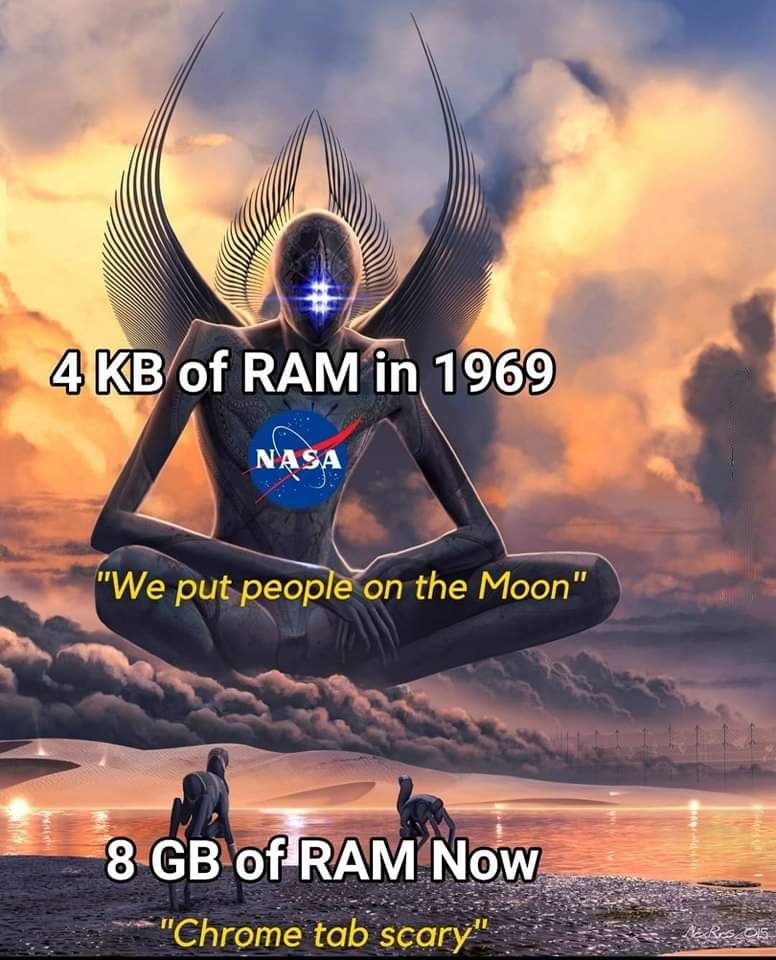}\end{minipage} &
\begin{minipage}[t]{0.58\textwidth}\vspace{0pt}
\raggedright
\{\newline
\hspace*{0.8em}\textcolor{blue}{\textbf{"visual"}}: \{\newline
\hspace*{1.6em}"description": "An illustration showing a giant, alien-like being sitting cross-legged in the sky, hovering over a landscape. Below, on the ground, two small figures look up at the giant being. Text is overlaid on the image in different positions. A circular blue logo with red and white elements is placed on the chest of the giant being.",\newline
\hspace*{1.6em}"ocr": "4 KB of RAM in 1969 / NASA / We put people on the Moon / 8 GB of RAM Now / Chrome tab scary",\newline
\hspace*{1.6em}"entities": ["e1: A giant, alien-like figure with glowing blue eyes and wings, sitting cross-legged in the sky.", "e2: Two small figures standing on the ground looking up.", "e3: A circular blue logo with a red chevron and white text NASA."]\newline
\hspace*{0.8em}\},\newline
\hspace*{0.8em}\textcolor{violet}{\textbf{"identity"}}: \{\newline
\hspace*{1.6em}"entities": ["e1: name: 4 KB of RAM in 1969; source: null.", "e2: name: 8 GB of RAM Now; source: null.", "e3: name: NASA; source: null."]\newline
\hspace*{0.8em}\},\newline
\hspace*{0.8em}\textcolor{orange}{\textbf{"knowledge"}}: \{\newline
\hspace*{1.6em}"facts": ["The Apollo Guidance Computer used during the Apollo missions had approximately 4 kilobytes of random-access memory (RAM).", "NASA's Apollo program successfully landed humans on the Moon for the first time in 1969 with the Apollo 11 mission.", "In contemporary consumer computing, 8 gigabytes of RAM is commonly regarded as an entry-level or baseline memory capacity for personal computers.", "Since the 2010s, Google Chrome has frequently been described by users and technology commentators as being resource-intensive, particularly in its use of system memory when multiple tabs or extensions are open."],\newline
\hspace*{1.6em}"cultural\_code": "Chrome RAM usage meme"\newline
\hspace*{0.8em}\},\newline
\hspace*{0.8em}\textcolor{green}{\textbf{"reasoning"}}: \{\newline
\hspace*{1.6em}"trigger": "The contrast between the text 4 KB of RAM in 1969 associated with the giant, powerful figure and the text 8 GB of RAM Now associated with the small, fearful figures.",\newline
\hspace*{1.6em}"reference": "The historical fact of the Apollo 11 moon landing versus the modern internet complaint about Google Chrome's memory consumption.",\newline
\hspace*{1.6em}"core\_message": "It is ironic that humanity accomplished the monumental feat of landing on the moon with only 4 KB of RAM, while modern computers with vastly more memory (8 GB) struggle to handle a single Google Chrome tab.",\newline
\hspace*{1.6em}"logic": "IR",\newline
\hspace*{1.6em}"logic\_detail": null\newline
\hspace*{0.8em}\}\newline
\}
\end{minipage} \\
\midrule
\begin{minipage}[t]{0.33\textwidth}\vspace{0pt}\centering
\includegraphics[width=\linewidth]{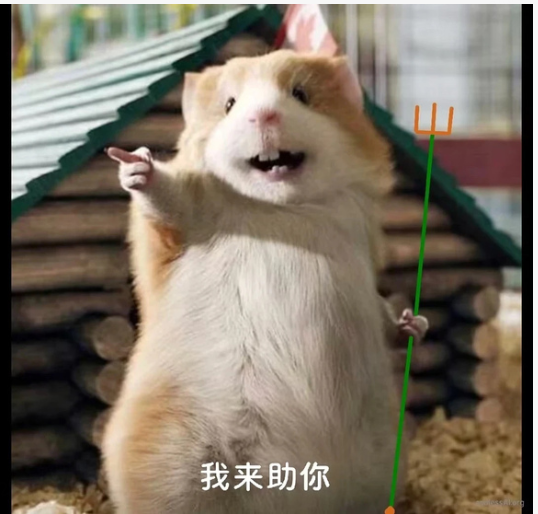}\par
\vspace{2pt}{\scriptsize\emph{Translation: I am here to help you.}\par}
\vspace{1pt}{\scriptsize\emph{Romanized: Xiyouji = Journey to the West.}\par}
\end{minipage} &
\begin{minipage}[t]{0.58\textwidth}\vspace{0pt}
\raggedright
\{\newline
\hspace*{0.8em}\textcolor{blue}{\textbf{"visual"}}: \{\newline
\hspace*{1.6em}"description": "A fluffy, anthropomorphic hamster standing on its hind legs in front of a small wooden log cabin. The hamster is pointing forward with its right paw and has its mouth open as if speaking or shouting. In its left paw, it holds a simple, poorly drawn green stick with an orange three-pronged fork top, resembling a trident. The text Wo Lai Zhu Ni is overlaid at the bottom.",\newline
\hspace*{1.6em}"ocr": "Wo Lai Zhu Ni",\newline
\hspace*{1.6em}"entities": ["e1: A standing hamster with an open mouth, pointing forward with one paw and holding a crudely drawn trident in the other."]\newline
\hspace*{0.8em}\},\newline
\hspace*{0.8em}\textcolor{violet}{\textbf{"identity"}}: \{\newline
\hspace*{1.6em}"entities": ["e1: name: Huangfeng Dasheng (Yellow Wind Sage); source: Heishenhua: Wukong (Black Myth: Wukong)."]\newline
\hspace*{0.8em}\},\newline
\hspace*{0.8em}\textcolor{orange}{\textbf{"knowledge"}}: \{\newline
\hspace*{1.6em}"facts": ["Heishenhua: Wukong is an action role-playing game developed by the Chinese game company Game Science and adapted from Xiyouji.", "Huangfeng Dasheng is a monster character in Heishenhua: Wukong. He resembles a rat-like demon and uses a trident as his weapon in battle.", "In Heishenhua: Wukong, Huangfeng Dasheng appears to assist the player during a fight against a giant beetle-like boss and shouts Wo Lai Zhu Ni.", "Wo Lai Zhu Ni is one of Huangfeng Dasheng's entrance lines in Heishenhua: Wukong and has been widely quoted among players because of its voice delivery and staging."],\newline
\hspace*{1.6em}"cultural\_code": "Wo Lai Zhu Ni (Black Myth: Wukong meme)"\newline
\hspace*{0.8em}\},\newline
\hspace*{0.8em}\textcolor{green}{\textbf{"reasoning"}}: \{\newline
\hspace*{1.6em}"trigger": "The combination of the cute hamster, the crudely drawn trident, and the exact quote Wo Lai Zhu Ni.",\newline
\hspace*{1.6em}"reference": "The epic entrance of the powerful Huangfeng Dasheng from the game Black Myth: Wukong.",\newline
\hspace*{1.6em}"core\_message": "The humor comes from the absurd contrast between a majestic, powerful game character's iconic entrance line and a cute, harmless hamster holding a badly drawn trident acting out the scene.",\newline
\hspace*{1.6em}"logic": "IC",\newline
\hspace*{1.6em}"logic\_detail": "Identity Collision (IC) mixed with Irony (IR). The meme maps a cute, harmless hamster holding a poorly drawn weapon to the powerful, fearsome rat demon from the game, creating a stark and humorous contrast between the grand, serious quote and the silly visual."\newline
\hspace*{0.8em}\}\newline
\}
\end{minipage} \\
\midrule
\begin{minipage}[t]{0.33\textwidth}\vspace{0pt}\centering
\includegraphics[width=\linewidth]{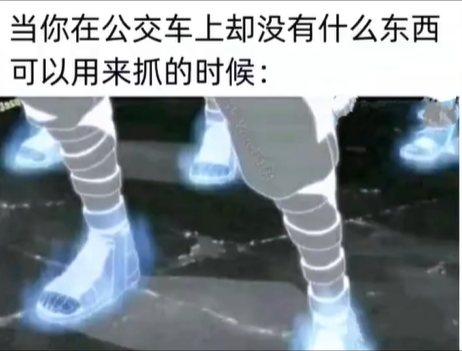}\par
\vspace{2pt}{\scriptsize\emph{Translation: When you are on a bus but have nothing to grab onto.}\par}
\vspace{1pt}{\scriptsize\emph{Romanized: Huoyingrenzhe = Naruto; chakela = chakra.}\par}
\end{minipage} &
\begin{minipage}[t]{0.58\textwidth}\vspace{0pt}
\raggedright
\{\newline
\hspace*{0.8em}\textcolor{blue}{\textbf{"visual"}}: \{\newline
\hspace*{1.6em}"description": "An image with text at the top and an illustration below. The text is in Chinese. The illustration shows a close-up of several pairs of feet wearing open-toed sandals and wrapped lower legs. The feet are glowing with a light blue aura against a dark, textured ground.",\newline
\hspace*{1.6em}"ocr": "When you are on a bus but have nothing to grab onto:",\newline
\hspace*{1.6em}"entities": ["e1: Feet wearing open-toed sandals and leg wraps, glowing with a blue aura."]\newline
\hspace*{0.8em}\},\newline
\hspace*{0.8em}\textcolor{violet}{\textbf{"identity"}}: \{\newline
\hspace*{1.6em}"entities": ["e1: name: Chakela Xifu; source: Huoyingrenzhe."]\newline
\hspace*{0.8em}\},\newline
\hspace*{0.8em}\textcolor{orange}{\textbf{"knowledge"}}: \{\newline
\hspace*{1.6em}"facts": ["In the world setting of the Japanese manga and anime Huoyingrenzhe, chakela is a supernatural energy formed by combining spiritual energy and physical energy.", "Chakela Xifu is a technique in Huoyingrenzhe in which ninja concentrate chakela on the soles of their feet or palms, allowing them to attach to surfaces such as trees, walls, and ceilings.", "By precisely controlling the amount and flow of chakela, ninja in Huoyingrenzhe can perform movements ordinary people cannot, such as walking on water or walking upside down."],\newline
\hspace*{1.6em}"cultural\_code": null\newline
\hspace*{0.8em}\},\newline
\hspace*{0.8em}\textcolor{green}{\textbf{"reasoning"}}: \{\newline
\hspace*{1.6em}"trigger": "The image of glowing feet from Huoyingrenzhe combined with the text about having nothing to hold onto on a bus.",\newline
\hspace*{1.6em}"reference": "The everyday struggle of keeping balance on a moving bus without a handrail, solved by the fictional ninja ability to magically stick feet to the floor.",\newline
\hspace*{1.6em}"core\_message": "It playfully suggests that one needs ninja-level magical abilities to maintain balance on a moving bus without holding a handrail.",\newline
\hspace*{1.6em}"logic": "IR",\newline
\hspace*{1.6em}"logic\_detail": "Applying a highly specialized, magical combat skill (sticking to walls with chakela) to a mundane, relatable daily annoyance (balancing on a bus)."\newline
\hspace*{0.8em}\}\newline
\}
\end{minipage} \\
\midrule
\begin{minipage}[t]{0.33\textwidth}\vspace{0pt}\centering
\includegraphics[width=\linewidth]{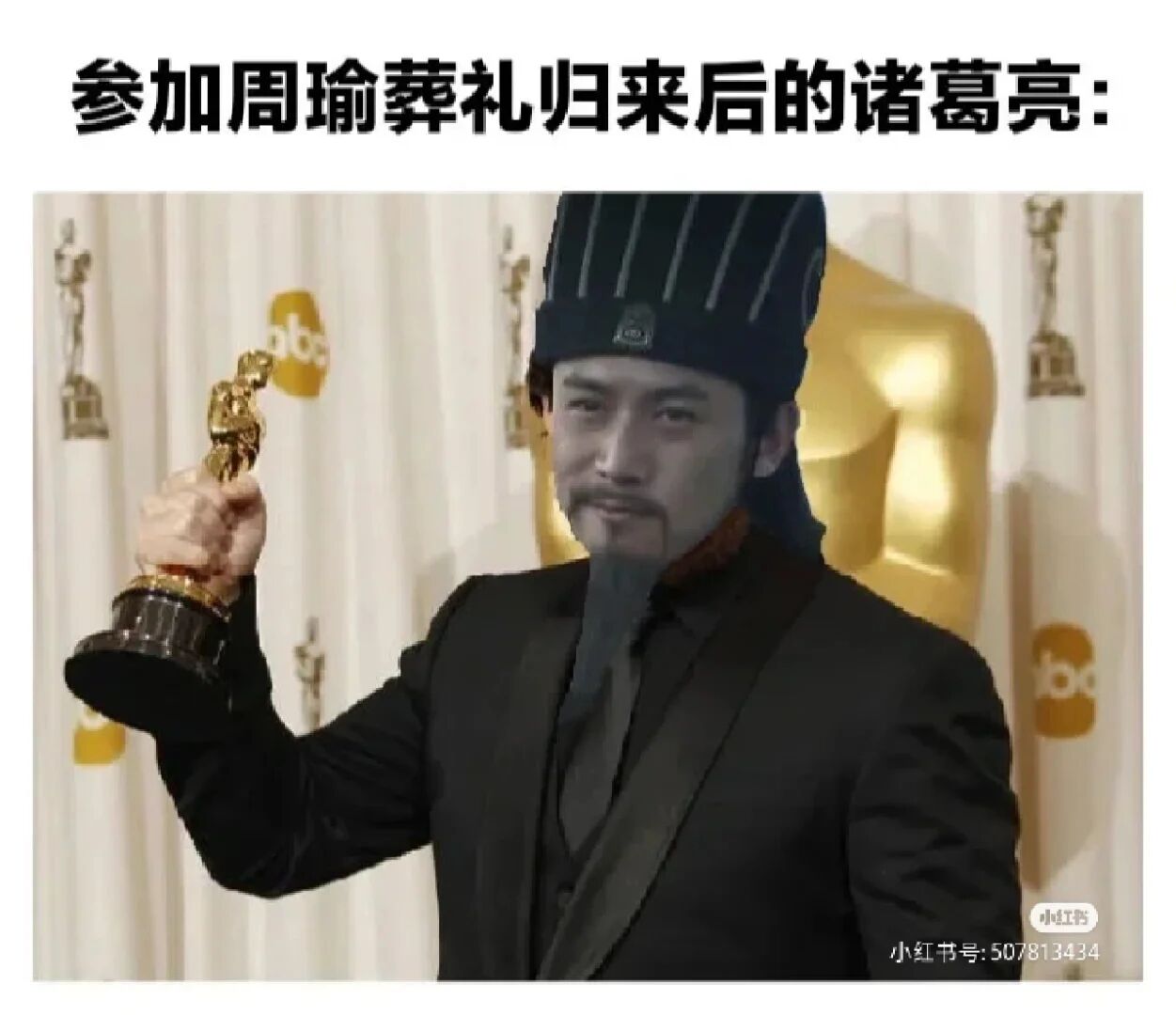}\par
\vspace{2pt}{\scriptsize\emph{Translation: Zhuge Liang after returning from Zhou Yu's funeral.}\par}
\vspace{1pt}{\scriptsize\emph{Romanized: Sanguo Yanyi = Romance of the Three Kingdoms; Diaoxiao = staged weeping.}\par}
\end{minipage} &
\begin{minipage}[t]{0.58\textwidth}\vspace{0pt}
\raggedright
\{\newline
\hspace*{0.8em}\textcolor{blue}{\textbf{"visual"}}: \{\newline
\hspace*{1.6em}"description": "The image shows a man wearing a traditional Chinese ancient-style black hat with vertical stripes and a long black beard, grafted onto a body wearing a modern black suit. He is holding a golden statuette (resembling an Oscar) in his raised right hand. He has a subtle smirk. The background features white curtains and large golden silhouettes of the same statuette. There is black Chinese text at the top of the image.",\newline
\hspace*{1.6em}"ocr": "Zhuge Liang after returning from Zhou Yu's funeral: / ab / bo / Xiaohongshu / Xiaohongshu ID: 507813434",\newline
\hspace*{1.6em}"entities": ["e1: A man's face with a long black beard and a traditional ancient Chinese black hat with vertical stripes, superimposed on a body wearing a modern black suit, holding a golden statuette.", "e2: A golden statuette of a standing figure, held by the man, with larger versions in the background."]\newline
\hspace*{0.8em}\},\newline
\hspace*{0.8em}\textcolor{violet}{\textbf{"identity"}}: \{\newline
\hspace*{1.6em}"entities": ["e1: name: Zhuge Liang; source: Sanguo Yanyi.", "e2: name: Oscar statuette (Academy Award); source: Academy Awards."]\newline
\hspace*{0.8em}\},\newline
\hspace*{0.8em}\textcolor{orange}{\textbf{"knowledge"}}: \{\newline
\hspace*{1.6em}"facts": ["In Sanguo Yanyi, Zhuge Liang and Zhou Yu have a complex relationship of rivalry and cooperation, and the two are portrayed as rivals who admire each other while competing against each other.", "Sanguo Yanyi includes a scene in which Zhuge Liang goes to mourn Zhou Yu, weeping bitterly and speaking with such sincerity that those present are deeply moved.", "Zhuge Liang Diaoxiao is often used in Chinese online culture and derivative works to discuss acting skills, disguised emotion, and similar topics, becoming a scene label with extended meaning.", "The trophy of the Academy Awards is commonly called the Oscar statuette, a gilded little golden figure and an important symbol of the award."],\newline
\hspace*{1.6em}"cultural\_code": "Zhuge Liang Diaoxiao / Oscar Yingdi"\newline
\hspace*{0.8em}\},\newline
\hspace*{0.8em}\textcolor{green}{\textbf{"reasoning"}}: \{\newline
\hspace*{1.6em}"trigger": "The text Zhuge Liang after returning from Zhou Yu's funeral combined with the image of him holding an Oscar statuette.",\newline
\hspace*{1.6em}"reference": "The cultural reinterpretation that Zhuge Liang's intense weeping at Zhou Yu's funeral was actually just top-tier acting rather than genuine sorrow.",\newline
\hspace*{1.6em}"core\_message": "Zhuge Liang's grief at Zhou Yu's funeral was so fake yet convincingly performed that he deserves an Oscar for Best Actor.",\newline
\hspace*{1.6em}"logic": "IC",\newline
\hspace*{1.6em}"logic\_detail": "Identity Collision: Treating historical/fictional political maneuvering as a modern acting performance worthy of a film award."\newline
\hspace*{0.8em}\}\newline
\}
\end{minipage} \\
\bottomrule
\end{longtable}
\endgroup

\end{document}